# Reconfiguration of a parallel kinematic manipulator with 2T2R motions for avoiding singularities through minimizing actuator forces

Francisco Valero [a], Miguel Díaz-Rodríguez [b], Marina Vallés [c], Antonio Besa [a], Enrique Bernabéu [c], Ángel Valera [c]

[a] Centro de Investigación en Ingeniería Mecánica, Universitat Politècnica de València, 46022, Spain.

[b] Departamento de Tecnología y Diseño, Facultad de Ingeniería, Universidad de los Andes, 5101, Mérida, Venezuela.

[c] Instituto Universitario de Automática e Informática Industrial, Universitat Politècnica de València, 46022, Spain.

Corresponding Author: Dr. Francisco Valero (fvalero@mcm.upv.es)

## Abstract

This paper aims to develop an approach for the reconfiguration of a parallel kinematic manipulator (PKM) with four degrees of freedom (DoF) designed to tackle tasks of diagnosis and rehabilitation in an injured knee. The original layout of the 4-DoF manipulator presents Type-II singular configurations within its workspace. Thus, we proposed to reconfigure the manipulator to avoid such singularities (owing to the Forward Jacobian of the PKM) during typical rehabilitation trajectories. We achieve the reconfiguration of the PKM through a minimization problem where the design variables correspond to the anchoring points of the robot limbs on fixed and mobile platforms. The objective function relies on the minimization of the forces exerted by the actuators for a specific trajectory. The minimization problem considers constraint equations to avoid Type-II singularities, which guarantee the feasibility of the active generalized coordinates for a particular path. To evaluate the proposed conceptual strategy, we build a prototype where reconfiguration occurs by moving the position of the anchoring points to holes bored in the fixed and mobile platforms. Simulations and experiments of several study cases enable testing the strategy performance. The results show that the reconfiguration strategy allows obtaining trajectories having minimum actuation forces without Type-II singularities.

# 1. INTRODUCTION

Rehabilitation robotics is a research field of growing interest. In this vein, authors of [3,4] proposed a novel 4 degrees-of-freedom (DoF) parallel kinematic mechanism (PKM) for lower limb rehabilitation (LLR) to aid in the rehabilitation and diagnosis (rehagnosis) of a knee joint. However, as many PKMs, especially those with lower mobility (less than 6 DoF), the proposed PKM presents forward kinematic singularities, or Type 2 singularities (Gosselin and Angeles [2]) which is among the worst problems a PKM can face [1]. On the other hand, the proposed mechanism has the advantage of being free of inverse kinematic singularities or Type 1. This is an important fact because the concept of developing a rehagnosis robot is that the therapist provides the required movements to the patient’s limb while the PKM records the trajectory required for the specific trajectory. In this case, since the PKM is free of inverse singularities, the PKM can perform and record the trajectory as long as the therapist provides the movement through the mobile platform. Having recorded the trajectory, the therapist switches to rehabilitation mode so that the PKM starts replicating the rehabilitation movements. In rehabilitation mode, the mechanism could lead to forward singularities, which means that the PKM gains at least one DoF and cannot resist some of the wrenches applied to its platform. Furthermore, the robot cannot pass through such a singular configuration without external help. The objective of this paper is to develop a strategy for dealing with Type 2 singularities of the 4 DoF proposed in [3,4].

Mainly, researchers have proposed two strategies to enhance the ability of a PKM to avoid singularities within the workspace: i) PKM with redundancy, and ii) Reconfigurable PKM (RPKM). A mechanism gains redundancy by introducing an extra active joint to a kinematics chain or by actuating one or more of its passive joints. See [26] for a more profound classification of redundant PKM.

On the other hand, an RPKM adopts different assembling configurations with different kinematic characteristics and dynamic behaviors by modifying passive joints, rigid links, mobile platforms, and end-effectors [5]. To our point of view, we can classify three main strategies for reconfiguration. In the first strategy, the robot can get another configuration through a modular design in which the components are building blocks allowing different assembling options. For instance, since a parallel mechanism has legs connected in parallel, it is possible to attach or detach the legs from the mobile platform to obtain a new configuration of the mechanism. In this regard, Xi et al. [6] propose a reconfigurable strategy based on two tripod PKMs in which the mobile and fixed platforms are connected in parallel. The proposed mechanism is able to reconfigure from 6 to 3 DoF. Reference [7-10] provides examples of building RPKMs using a modular design.

The second strategy of RPKMs takes advantage of lockable joints. For instance, in Carbonari et al. [11] the design of the spherical joints relies on a combination of revolute joints, which allows the mechanism to take different kinematic configurations with different types of mobility by alternately fixing some of the revolute joints. References [12-14] have examples of other lockable joint reconfiguration strategies.

A third approach to developing an RPKM is by adjusting the link size/orientation or modifying the size of the mobile or fixed platform. We refer to this group of RPKMs as a variable geometric reconfiguration that makes it possible to increase the workspace or avoid singularities without modifying either the robot topology or the end-effector's DoF. The mechanisms in this group have a kinematics chain providing reconfiguration capabilities to the mechanism. Coppola et al. [15] introduced a bevel gear system to modify simultaneously the position of the first revolute (R) joint of each chain, thus modifying the size of the fixed platform of an RRPS (revolute-revolute-prismatic-spherical). References [16-21] provide examples of RPKM following this approach. In this strategy, we can include those PKM that reconfigure by introducing one or several extra kinematic chains provided that this extra chain keeps locked when the mechanism follows the trajectory, see kinematic redundancy mechanism [27, 28].

The main goal of this paper is to propose an approach for allowing the PKM proposed by Araujo-in [3] and [4] to follow the trajectory avoiding singular configuration while keeping the actuator forces to a minimum. In this vein, we keep the mechanical complexity of the system as low as possible, by considering a reconfiguration strategy of the first type. We follow a modular design providing the mobile and fixed platforms the ability to reconfigure by changing the anchoring points in both platforms. Thus, in this paper, we consider the reconfiguration of the PKM as an optimization problem where the design variables correspond to the anchoring points of the mechanism limbs on both fixed and mobile platforms. We propose to minimize the forces supplied by the actuators during a specific trajectory avoiding forward singularities to guarantee the feasibility of the active generalized coordinates. Note that [27-29] and reference therein develop analogous strategies to the one proposed in this paper, yet based on the kinematic redundancy concept.

Section 2 presents the kinematic model of PKMs and the forward singularity condition equation. In section 3, we develop the dynamic model of the PKM. Section 4, summarizes the design procedure for the RPKM relaying on a non-linear optimization problem subject to both linear and non-linear constraints. Section 5 presents the implementation of the proposed approach. Section 6, shows the numerical evaluation applied to a virtual RPKM. Section 7, presents the experimental evaluation of the actual RPKM. Section 8 summarizes the conclusions.

# 2. KINEMATIC MODEL

The design specifications for the parallel robot can be established according to the analysis of the basic movements required for rehabilitation and considering the type of tests used for diagnosis. For rehabilitation purposes, different exercises can be performed with parallel robots: passive (without any voluntary movement by the patient) or active (with voluntary movements) exercises.

For diagnostic purposes, two main diagnostic tests are performed after knee surgery. The first one is the Lachman test, which assesses anterior cruciate ligament tear by displacing the tibia relative to the femur. The second one is the pivot shift test, which is intended to reproduce translational and rotational instability in the knee by applying a twist to the tibia and essentially measuring the rotation.

In previous papers, [3] and [4], the design and kinematic analysis of a 3UPS/RPU PKM with 2T2R motion for knee diagnosis and rehabilitation tasks has been presented. The considered PKM has three identical external limbs composed of universal, prismatic (actuated), and universal joints. In addition, the PKM presents a central strut with rotational, prismatic (actuated), and universal joints. The 3UPS-RPU PKM is depicted in Figure 1.

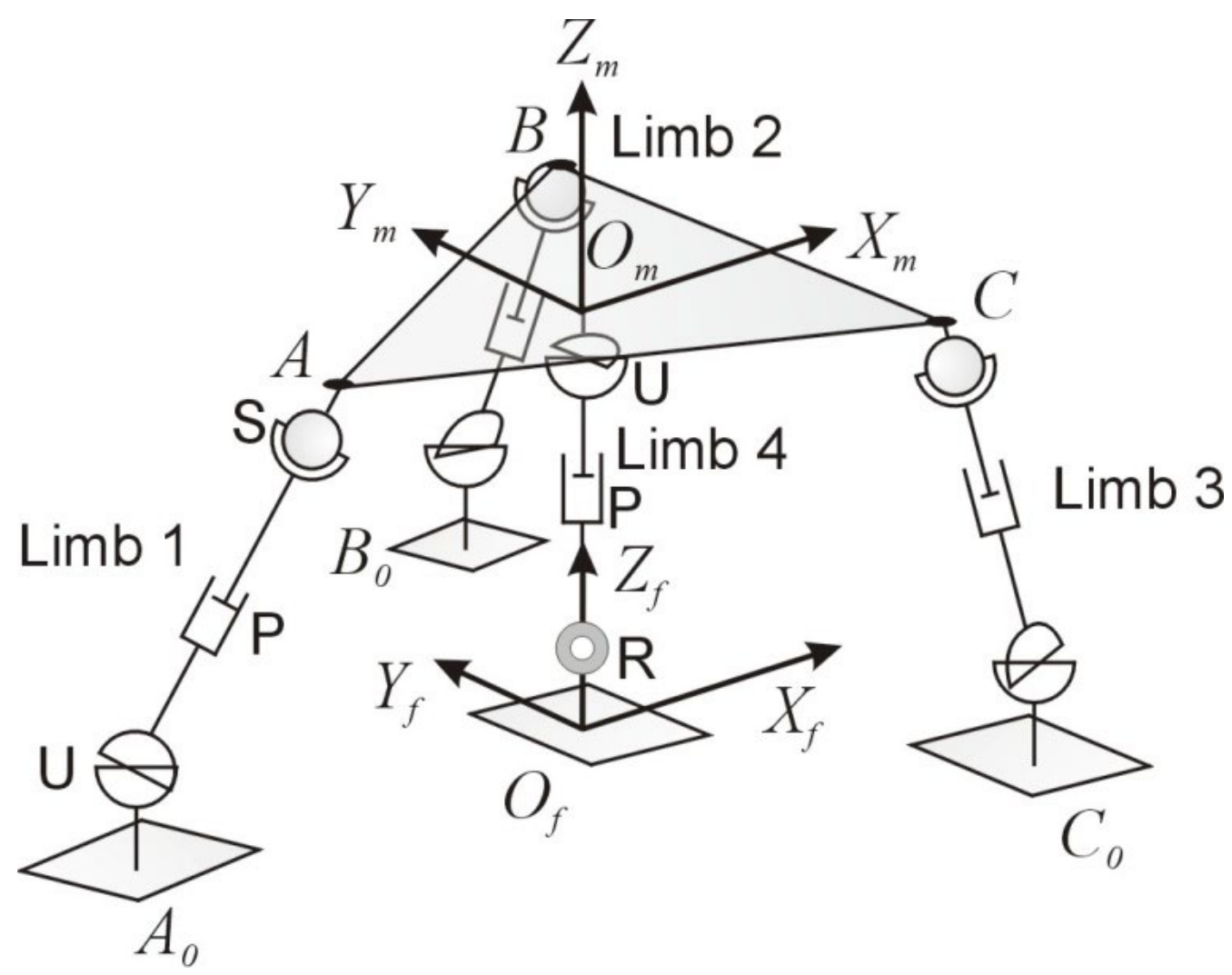

Figure 1. The 3UPS-RPU parallel manipulator

Note that in Figure 1 the fixed reference system is given by $\{O_f - X_f Y_f Z_f\}$ and the reference system attached to the mobile platform is denoted by $\{O_m - X_m Y_m Z_m\}$. This PKM, with four degrees of freedom, provides the intended motion for human body lower extremity rehabilitation tasks; namely, two translations within the $X_f Z_f$ plane, which correspond to the sagittal plane of the patient, and two rotations, one around the $Y_m$ axis and the other one around the $Z_m$ axis, both corresponding to the system of the reference attached to the mobile platform. Notice that has been choose the mobile reference system instead of the fixed one, due that some sensors will be located on the mobile platform and also considering the mechanical capabilities of the universal joint of the central limb. For the sake of completeness, the kinematic problem is briefly described below. A detailed description of the kinematic modeling of the robot can be found in [4].

Considering the Denavit-Hartenberg (DH) notation, Asada y Slotine [30], this manipulator could be modeled through a set of 22 generalized dependent coordinates $q_{ij}$, where the first subscript corresponds to the number of the limb and the second to the coordinate within the limb. The corresponding DH parameters for the external UPS limbs and the central RPU limb are depicted in Tables 1 and 2, respectively. In Figure 2, the generalized coordinates are depicted for both the external limbs and the central ones.

Table 1. DH parameters for *i*-th UPS external limb

| $\alpha_i$ | $a_i$ | $d_i$ | $\theta_i$ |
|---|---|---|---|
| $-\frac{\pi}{2}$ | $0$ | $0$ | $q_{i1}$ |
| $\frac{\pi}{2}$ | $0$ | $0$ | $q_{i2}$ |
| $0$ | $0$ | $q_3$ | $0$ |
| $\frac{\pi}{2}$ | $0$ | $0$ | $q_{i4}$ |
| $\frac{\pi}{2}$ | $0$ | $0$ | $q_{i5}$ |
| $\frac{\pi}{2}$ | $0$ | $0$ | $q_{i6}$ |

Table 2. DH parameters for the 4 RPU central limb

| $\alpha_i$ | $a_i$ | $d_i$ | $\theta_i$ |
|---|---|---|---|
| $-\frac{\pi}{2}$ | $0$ | $0$ | $q_{41}$ |
| $-\frac{\pi}{2}$ | $0$ | $q_2$ | $\pi$ |
| $+\frac{\pi}{2}$ | $0$ | $0$ | $q_{43}$ |
| $0$ | $0$ | $0$ | $q_{44}$ |

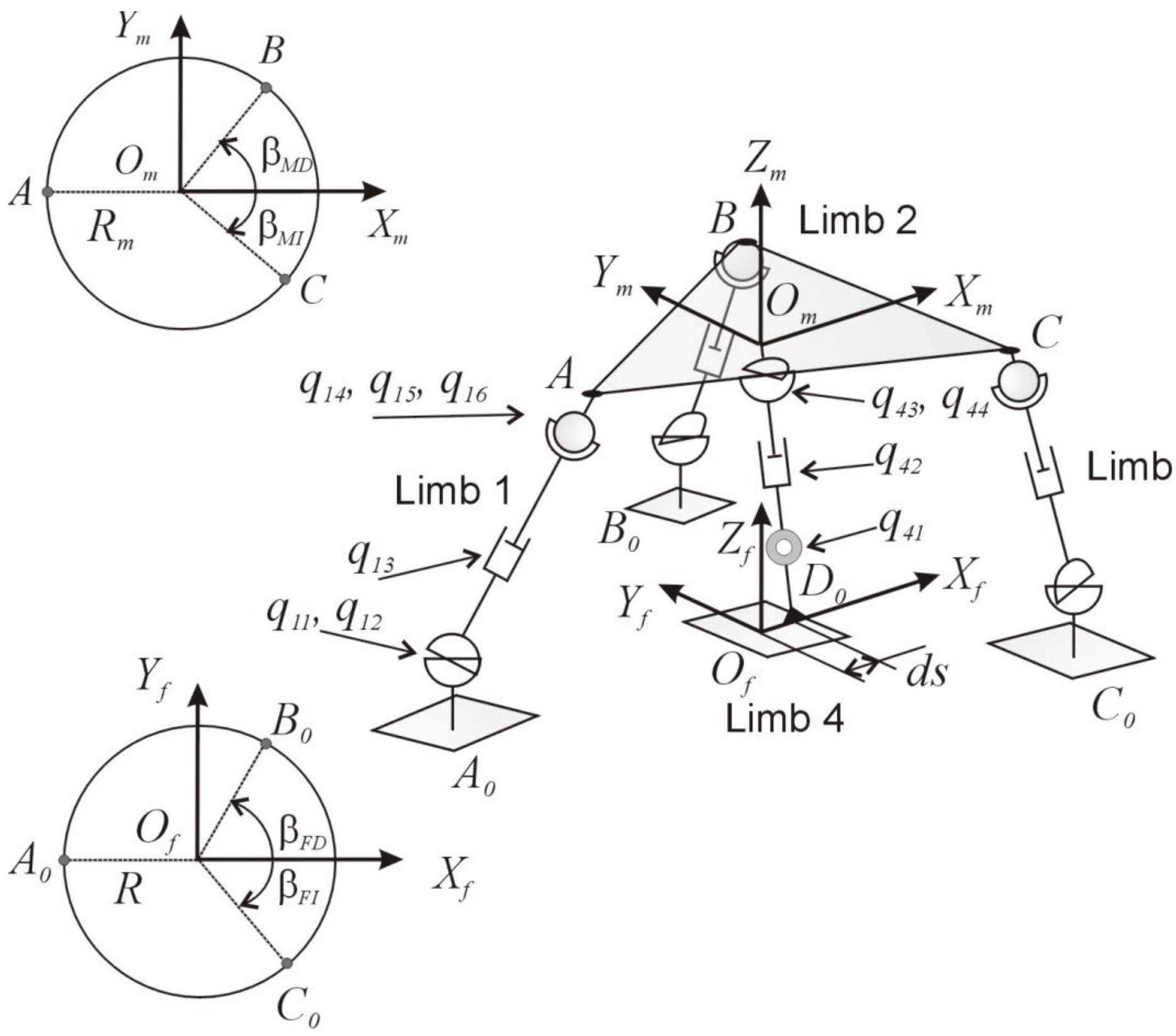

Figure 2. Kinematic modeling of the PKM

The active generalized coordinates will be $q_{13}$, $q_{23}$, $q_{33}$, and $q_{42}$, which correspond to the actuated prismatic joints. Also, the connection points of the four limbs with both the fixed and the mobile platform are depicted in Figure 2. The location of those points depends on the radius $R$ and the angles $\beta_{FD}$ and $\beta_{FI}$ for the fixed platform and $R_m$, $\beta_{MD,}$ and $\beta_{MI}$ for the mobile one. It should also be noted that the lower connection point of the central limb, $D_0$, can be located along the $X_f$ axis of the fixed reference system, $ds$ being the magnitude of that displacement. Some of these seven geometrical parameters ($R,\ \beta_{FD},\ \beta_{FI},\ R_m,\ \beta_{MD},\ \beta_{MI,}$ and $ds$) will be considered as the variables upon which the geometric reconfiguration of the robot will be optimized.

Notice that $x_m$ and $z_m$ are the coordinates of the origin of the mobile reference system ($y_m$ is always zero due to the robot topology) and $\theta$ and $\psi$ are the angles rotated by the mobile platform around $Y_m$ and $Z_m$, respectively. It is possible to obtain explicit expressions for the solution of the inverse kinematic problem relating these variables and the actuated generalized coordinates $q_{13}$, $q_{23}$, $q_{13}$ and $q_{42}$ as follows,

$$q_{13} = +\sqrt{\begin{array}{l} R^2 + \left(2 \cdot x_m - 2 \cdot C_\theta \cdot C_\psi \cdot R_m\right) \cdot R + R_m^2 + \\ + \left(2 \cdot z_m \cdot C_\psi \cdot S_\theta - 2 \cdot C_\theta \cdot C_\psi \cdot x_m\right) \cdot R_m + x_m^2 + z_m^2 \end{array}}$$

$$q_{23} = +\sqrt{\begin{array}{l} R^2 + R_m^2 + x_m^2 + z_m^2 + \\ + 2 \cdot R_m \cdot \begin{pmatrix} R \cdot \left(C_{FD} \cdot C_\theta \cdot S_{MD} \cdot S_\psi - C_{FD} \cdot C_{MD} \cdot C_\psi \cdot C_\theta - C_{MD} \cdot S_{FD} \cdot S_\psi - C_\psi \cdot S_{FD} \cdot S_{MD}\right) + \\ + C_\theta \cdot C_\psi \cdot C_{MD} \cdot x_m - C_{MD} \cdot C_\psi \cdot S_\theta \cdot z_m - C_\theta \cdot S_{MD} \cdot S_\psi \cdot x_m + S_\theta \cdot S_\psi \cdot S_{MD} \cdot z_m \end{pmatrix} - \\ - 2 \cdot R \cdot x_m \cdot C_{FD} \end{array}}$$

$$q_{33} = +\sqrt{\begin{array}{l} R^2 + R_m^2 + x_m^2 + z_m^2 + \\ + 2 \cdot R_m \cdot \begin{pmatrix} R \cdot \left(- C_{FI} \cdot C_\theta \cdot S_{MI} \cdot S_\psi - C_{FI} \cdot C_{MI} \cdot C_\psi \cdot C_\theta + C_{MI} \cdot S_{FI} \cdot S_\psi - C_\psi \cdot S_{FI} \cdot S_{MI}\right) + \\ + C_\theta \cdot C_\psi \cdot C_{MI} \cdot x_m - C_{MI} \cdot C_\psi \cdot S_\theta \cdot z_m + C_\theta \cdot S_{MI} \cdot S_\psi \cdot x_m - S_\theta \cdot S_\psi \cdot S_{MI} \cdot z_m \end{pmatrix} - \\ - 2 \cdot R \cdot x_m \cdot C_{FI} \end{array}}$$

$$q_{42} = +\sqrt{d_s^2 - 2 \cdot ds \cdot x_m + x_m^2 + z_m^2}$$

(1)

where $C_\theta$, $S_\theta$, $C_{FD}$, $S_{FD}$,… stand for $cos(\theta)$, $sin(\theta)$, $cos(\beta_{FD})$, $sin(\beta_{FD})$,…. Considering now the time derivatives of the equations (1), the following matrix expression relating the actuated generalized velocities and the velocities of the mobile platform, could be obtained,

$$\Phi_a \cdot \begin{bmatrix} \dot{q}_{13} \\ \dot{q}_{23} \\ \dot{q}_{33} \\ \dot{q}_{42} \end{bmatrix} = \Phi_x \cdot \begin{bmatrix} \dot{x}_m \\ \dot{z}_m \\ \dot{\theta} \\ \dot{\psi} \end{bmatrix} \qquad (2)$$

where $\Phi_a$ is the Inverse Jacobian, in this case, the identity matrix, so that no inverse singularities will appear in this PMK. However, the Forward Jacobian, $\Phi_x$, is a function of the mobile platform variables, $x_m$, $z_m$, $\theta$, $\psi$, and its determinant could become zero, which implies that the PKM is going through a forward singularity.

# 3. DYNAMIC MODEL

Considering that the movements to be carried out during the rehabilitation process will be at very low velocity, inertial forces will be excluded from the dynamic model. In the Figure 3, a scheme of the external forces taken into account is depicted, where $\vec{P}_{i1}, \vec{P}_{i2} \quad i = 1,..,4$ are the weights considered for the cylinder and the rod of each prismatic actuator, $\vec{P}_m$ is the weight of the mobile platform, $\vec{F}_m, \vec{T}_m$ are the external forces applied by the patient to the mobile platform, and $\vec{F}_{act_i} \quad i = 1,..,4$ are the active forces applied by the actuators. Regarding the friction, only the one produced in the prismatic actuators will be considered, $\vec{F}_{f_i} \quad i = 1,..,4$. The values of the weights and the location of the centers of gravity have been made from a CAD model of the PKM. So, the dynamic model of the PKM can be found without considering the generalized coordinates for the spherical and universal joint connecting the four limbs to the mobile platform. The following subset of 15 generalized dependent coordinates will be considered,

$$\vec{q} = \Big[\underbrace{q_{11}, q_{12}, q_{21}, q_{22}, q_{31}, q_{32}, q_{41}, x_m, z_m, \theta, \psi,}_{Secondary} \quad \underbrace{q_{13}, q_{23}, q_{33}, q_{42}}_{Independent}\Big]^T \tag{3}$$

where a partition has been introduced between the active (independent) and other (secondary) coordinates. The first step to obtain the equation of motion of the PKM will be to map the gravitational, external, friction, and applied forces into the space of generalized coordinates previously introduced. This could be done, for instance, using the Principle of Virtual Power. Let be one of the four limbs of the PKM, the prismatic actuator is decomposed in two bars, Figure 3, where the $G_{i1}$, $G_{i2}$ are the center of mass of each bar, and $\vec{v}_{G_{i1}}$ and $\vec{v}_{G_{i2}}$ the velocities of both about the fixed reference system. Similarly $G_m$ is the center of mass of the mobile platform and $\vec{v}_{G_m}$ its velocity. The gravitational generalized forces will be,

$$\vec{Q}_{grav} = \sum_{i=1}^{4} \vec{P}_{i1} \cdot \frac{\partial \vec{v}_{G_{i1}}}{\partial \dot{\vec{q}}} + \sum_{i=1}^{4} \vec{P}_{i2} \cdot \frac{\partial \vec{v}_{G_{i2}}}{\partial \dot{\vec{q}}} + \vec{P}_m \cdot \frac{\partial \vec{v}_{G_m}}{\partial \dot{\vec{q}}} \tag{4}$$

where $\dot{\vec{q}}$ is the vector of generalized velocities done by,

$$\dot{\vec{q}} = \left[\dot{q}_{11}, \dot{q}_{12}, \dot{q}_{21}, \dot{q}_{22}, \dot{q}_{31}, \dot{q}_{32}, \dot{q}_{41}, \dot{x}_m, \dot{z}_m, \dot{\theta}, \dot{\psi}, \dot{q}_{13}, \dot{q}_{23}, \dot{q}_{33}, \dot{q}_{42}\right]^T \tag{5}$$

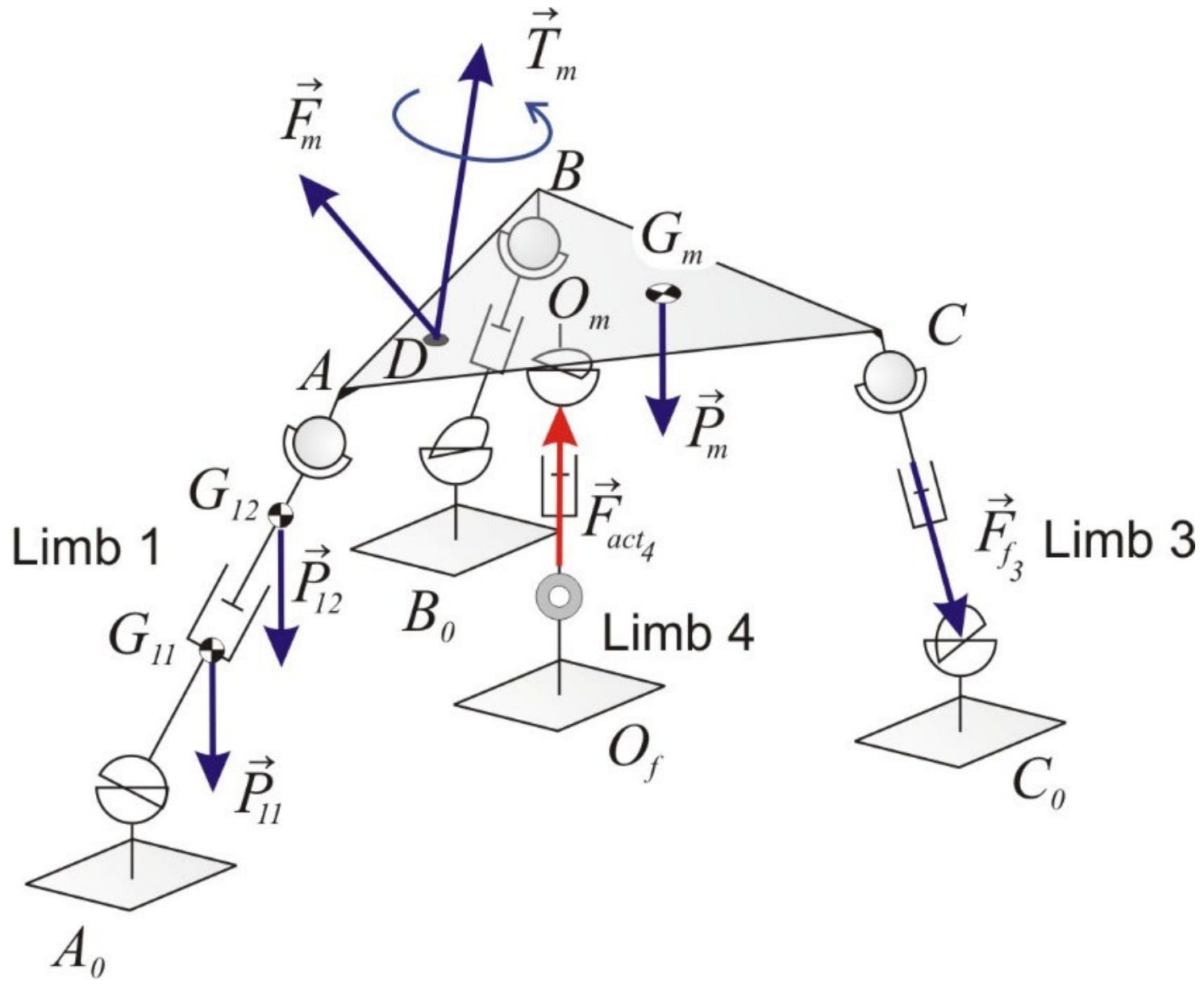

Figure 3. External forces considered in the dynamic model

The generalized forces corresponding to the external forces applied by the patient to the mobile platform will be obtained as follows,

$$\vec{Q}_{ext} = \vec{F}_m \cdot \frac{\partial \vec{v}_D}{\partial \dot{\vec{q}}} + \vec{T}_m \cdot \frac{\partial \vec{\omega}_m}{\partial \dot{\vec{q}}} \tag{6}$$

where $\vec{v}_D$ is the point of application of external forces and $\vec{\omega}_m$ the angular velocity of the mobile platform, both kinematic terms about the fixed reference system. The generalized forces corresponding to the active forces will be obtained as follows,

$$\vec{Q}_{act} = \sum_{i=1}^{4} \vec{F}_{act_i} \cdot \frac{\partial \vec{v}_{G_{i2}}}{\partial \dot{\vec{q}}} \tag{7}$$

where $\vec{F}_{act_i}$ is the force applied by the *i*-th prismatic actuator. In a similar way, the generalized friction force will be,

$$\vec{Q}_{fric} = \sum_{i=1}^{4} \vec{F}_{f_i} \cdot \frac{\partial \vec{v}_{G_{i2}}}{\partial \dot{\vec{q}}} \tag{8}$$

where $\vec{F}_{f_i}$ is the friction force considered in the i-th prismatic joint. The modulus of this force will be,

$$F_{f_i} = -sign(\dot{q}_{i3}) \cdot (\mu_c + \mu_v \cdot |\dot{q}_{i3}|) \tag{9}$$

$\mu_c$ and $\mu_v$ being the Coulomb and viscous coefficients, respectively. The friction in the four actuators has been experimentally identified. In Figure 4, the values corresponding to the friction in the central strut, the most robust of the four are depicted. There is good agreement between the theoretical model and the experimental results.

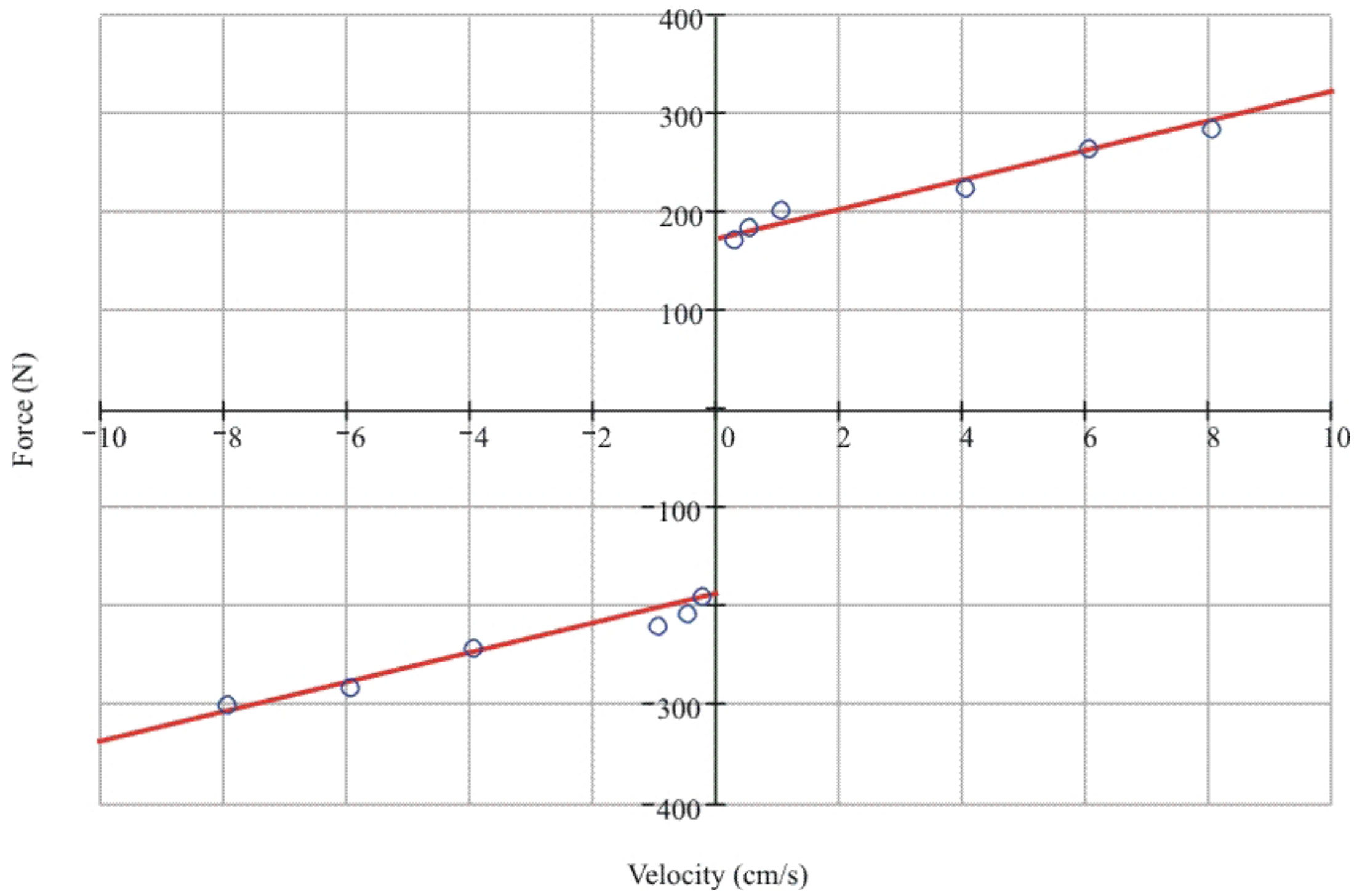

Figure 4. Friction force in the central limb

Taking into account that the PKM has been modelled using a set of dependent coordinates, the equation of motion of the mechanical system can be written as follows,

$$\vec{Q}_{grav} + \vec{Q}_{ext} + \vec{Q}_{act} + \vec{Q}_{fric} + \Phi_q^T \cdot \vec{\lambda} = 0 \tag{10}$$

where $\vec{Q}_{grav}, \vec{Q}_{ext}, \vec{Q}_{fric}, \vec{Q}_{act} \in R^{15}$. $\Phi_q$ is the Jacobian given by,

$$\Phi_q = \left[ \frac{\partial \vec{\Phi}}{\partial \vec{q}} \right] \tag{11}$$

being $\vec{\Phi}$ the vector grouping the following 11 constraint equations,

$$\begin{aligned}
&\Phi_1 \equiv C_{11} \cdot S_{12} \cdot q_{13} - R - x_m + C_\theta \cdot C_\psi \cdot R_m \\
&\Phi_2 \equiv -C_{12} \cdot q_{13} + S_\psi \cdot R_m \\
&\Phi_3 \equiv S_{11} \cdot S_{12} \cdot q_{13} - z_m - S_\theta \cdot C_\psi \cdot R_m \\
&\Phi_4 \equiv C_{21} \cdot S_{22} \cdot q_{23} + R \cdot C_{FD} - x_m - C_\theta \cdot C_\psi \cdot C_{MD} \cdot R_m + C_\theta \cdot S_\psi \cdot S_{MD} \cdot R_m \\
&\Phi_5 \equiv -C_{22} \cdot q_{23} + R \cdot S_{FD} - s_\psi \cdot C_{MD} \cdot R_m - C_\psi \cdot S_{MD} \cdot R_m \\
&\Phi_6 \equiv S_{21} \cdot S_{22} \cdot q_{23} - z_m - S_\theta \cdot C_\psi \cdot C_{MD} \cdot R_m - S_\theta \cdot S_\psi \cdot S_{MD} \cdot R_m \\
&\Phi_7 \equiv C_{31} \cdot S_{32} \cdot q_{33} + R \cdot C_{FI} - x_m - C_\theta \cdot C_\psi \cdot C_{MI} \cdot R_m - C_\theta \cdot S_\psi \cdot S_{MI} \cdot R_m \\
&\Phi_8 \equiv -C_{32} \cdot q_{33} - R \cdot S_{FI} - s_\psi \cdot C_{MI} \cdot R_m + C_\psi \cdot S_{MI} \cdot R_m \\
&\Phi_9 \equiv S_{31} \cdot S_{32} \cdot q_{33} - z_m + S_\theta \cdot C_\psi \cdot C_{MI} \cdot R_m + S_\theta \cdot S_\psi \cdot S_{MI} \cdot R_m \\
&\Phi_{10} \equiv -S_{41} \cdot q_{42} + ds - x_m \\
&\Phi_{11} \equiv C_{41} \cdot q_{42} - z_m
\end{aligned} \tag{12}$$

and $\vec{\lambda} \in R^{11}$ is the vector of Lagrange unknown multipliers. Notice that the term $\Phi_q^T \cdot \vec{\lambda} = 0$ stands for the internal generalized forces.

Of the various procedures proposed to eliminate the internal generalized forces, the coordinate partitioning procedure has been selected. The matrix relating generalized velocities to those chosen as independent will be,

$$\begin{bmatrix} \dot{\vec{q}}^s \\ \dot{\vec{q}}^i \end{bmatrix} = \begin{bmatrix} -\left(\Phi_q^s\right)^{-1} \cdot \Phi_q^i \\ I \end{bmatrix} \cdot \left[ \dot{\vec{q}}^i \right] \tag{13}$$

where $\Phi_q^i, \Phi_q^s$ are the part of the Jacobian matrix corresponding to the independent and secondary generalized coordinates, respectively, and $I$ is a 3x3 identity matrix, so that,

$$R^* = \begin{bmatrix} -\left(\Phi_q^s\right)^{-1} \cdot \Phi_q^i \\ I \end{bmatrix}_{N \times F} \tag{14}$$

where $N$ (15) is the number of generalized coordinates considered in the dynamic model, and $F$ (4) is the degrees of freedom of the PKM. It is important to point out that the determinants of both Jacobian matrices, $\Phi_q^s$, and $\Phi_x$ vanish in the same manipulator configurations, the latter being of significantly smaller dimensions. Therefore, as is well known, by preventing the Jacobian determinant from becoming null, not only are the control problems associated with direct singularities avoided, but disproportionately large control actions can also be prevented, for reasonable conditions of movement, in the vicinity of the singular configurations. This problem will be illustrated in the following Section.

The equation of motion can be put in a more compact form as follows,

$$\left(R^*\right)^T_{F\times N}\cdot\left(\vec{Q}_{grav_{N\times1}}+\vec{Q}_{ext_{N\times1}}+\vec{Q}_{act_{N\times1}}+\vec{Q}_{fric_{N\times1}}\right)=\vec{0}_{F\times1} \quad (15)$$

from this equation, the active forces that the actuators have to apply to the PKM can be easily obtained.

# 4. OPTIMIZATION PROCEDURE

The objective pursued with the optimization is to facilitate the reconfiguration of the geometry of the robot so that it is feasible to perform each of the paths specified by the rehabilitation specialists, for this the singularities must be avoided and the physical limitations of the mechanical components must be considered, also seeking that the forces required of the actuators be reduced. It is also intended to make the reconfiguration of the robot a simple task so that the number of geometric parameters to be modified is not excessive. The problem is addressed in two steps; in the first one, a large number of parameters is assumed, given the results in the second, the parameters associated with the mobile platform are set and the rest is worked on.

The proposed reconfiguration procedure is based on the modification of a set of geometric parameters of the robot, distinguishing between those associated with the mobile platform, $PR_M=\{R_m,\beta_{MD},\beta_{MI}\}$, and those associated with the fixed one, $PR_F=\{R,ds,\beta_{FD},\beta_{FI}\}$ (see

Figure 5). $R$ and $R_m$ are the radii of the circumferences on which the robot limbs are attached to the fixed and mobile platforms, *ds* is the distance from the origin of the fixed reference system to the base of the central limb, $\beta_{FD}$ and $\beta_{FI}$ are the angles that define the positions of the lateral limbs on the fixed base, and $\beta_{MD}$ and $\beta_{MI}$ are the angles corresponding to the position of the lateral limbs on the mobile platform.

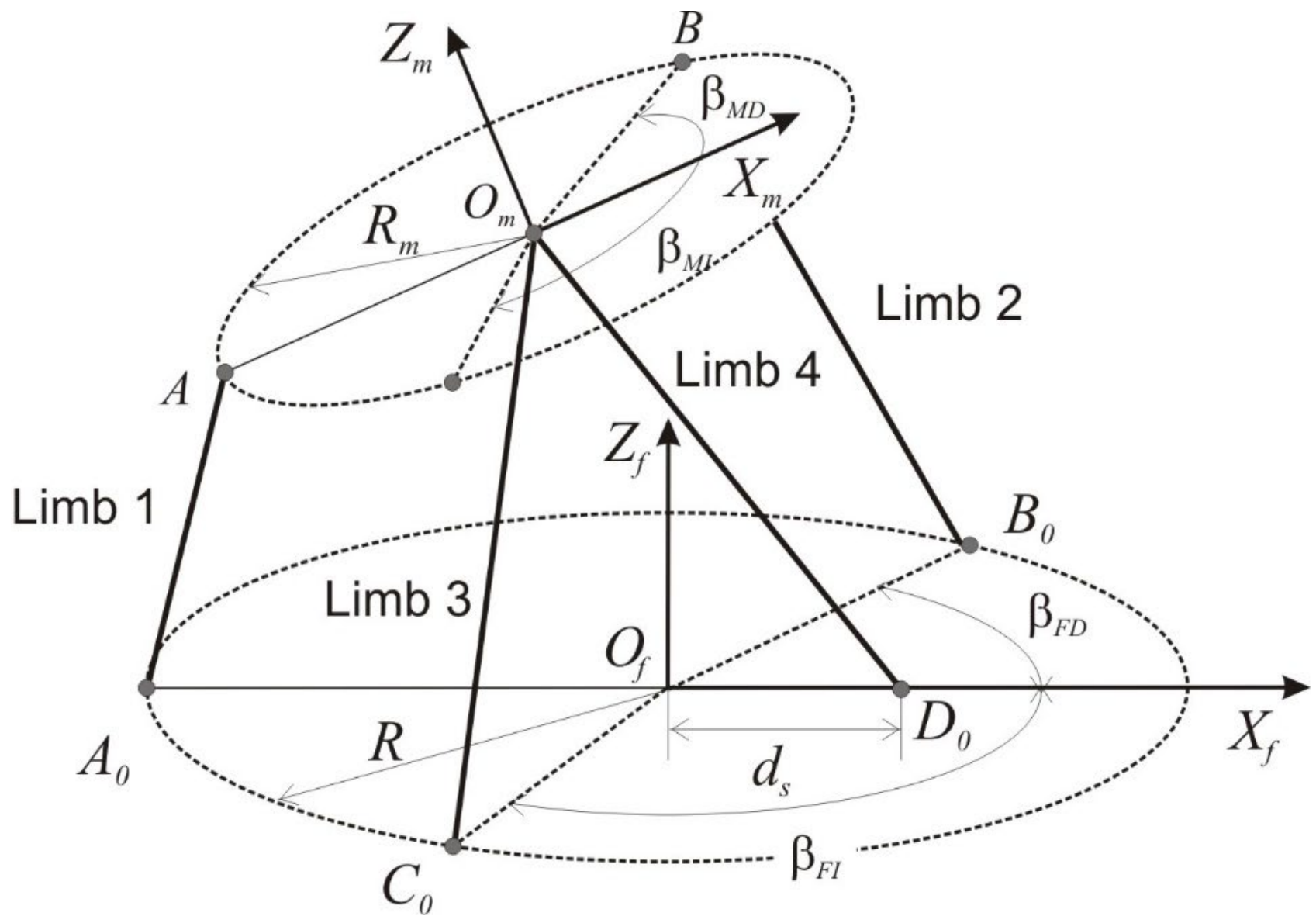

Figure 5. Schema of the reconfiguration variables.

The optimization process is approached in two stages. In the first one, the 7 variables corresponding to the mobile and fixed platforms will be considered, $PR_{7v} = PR_M \cup PR_F$, corresponding to the complete set of parameters, and for each *j*-th trajectory ($Tr_j$) the optimization procedure described in Subsection 4.1 will be applied in order to obtain a set of parameters $PR^j_{7v} = PR^j_M \cup PR^j_F$. In the second stage, to simplify the operation of the robot reconfiguration, only the variables associated with the anchors of the actuators on the fixed platform will be considered, $PR^j_{4v} = PR^j_F$, so that the parameters associated with the mobile platform will be fixed, making them equal to the median of the values obtained in phase one, $PR^c_M = Median\{PR_M^{j=1,..n}\}$, and the optimization procedure detailed in Subsection 4.2 will be applied in order to obtain the results $PR^j_{4v}$ for the *n* trajectories (*j* =1,..*n*).

So, at the end of the optimization procedure we will have the robot reconfigurations with 4 and 7 variables ($PR_{7v}^{j}$ and $PR_{4v}^{j}$) for each trajectory $j$. Figure 6 shows a flowchart of the proposed optimization approach.

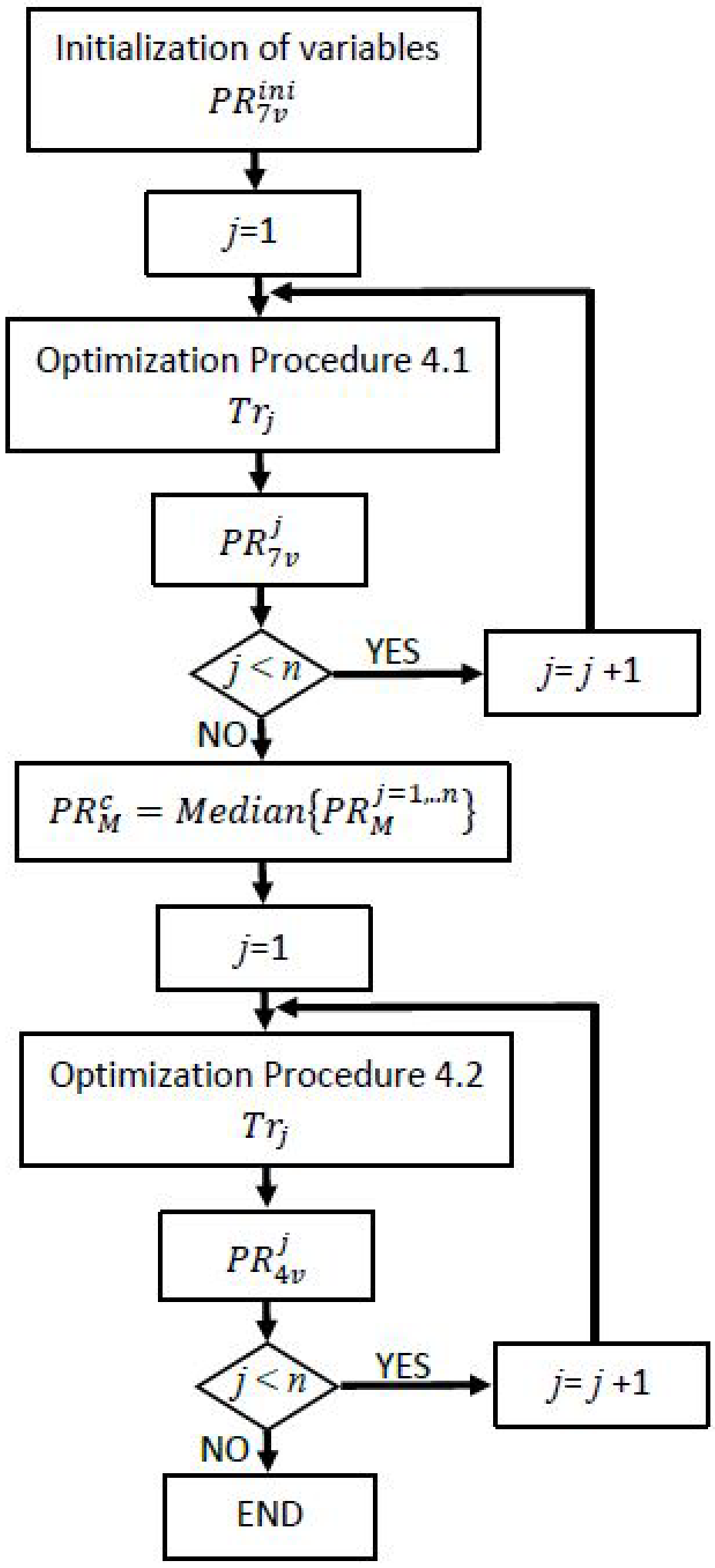

Figure 6. Flow diagram of the optimization procedure

### 4.1. Optimization procedure with seven variables

The following 7 design variables are selected for the optimization procedure (see Figure 5): radius of the fixed and mobile platforms, $R$ and $R_m$, distance $ds$, and angles $\beta_{FD}, \beta_{FI}$ for the base of the robot and $\beta_{MD}, \beta_{MI}$ for the mobile one. With criteria based on the final size of the assembly and geometric compatibility, the following limit values for the parameters are established,

$$\left.\begin{array}{lll} R & \in & [0.20, 0.50]m \\ R_m & \in & [0.15, 0.30]m \\ ds & \in & [-0.15, 0.15]m \\ \beta_{FD} & \in & [0.10, \pi]rad \\ \beta_{FI} & \in & [0.10, \pi]rad \\ \beta_{MD} & \in & [0.10, \pi]rad \\ \beta_{MI} & \in & [0.10, \pi]rad \end{array}\right\} \quad (16)$$

The trajectory followed by the PKM will be discretized into a sufficient number, $p$, of via points. The objective function will be computed at these points and the corresponding constraint equations will be considered. The objective function to minimize will be the sum of the square of the four active generalized forces corresponding to each of the configurations considered and obtained through the inverse dynamics of the RPKM,

$$f(R, R_m, ds, \beta_{FD}, \beta_{FI}, \beta_{MD}, \beta_{MI}) = \sum_{i=1}^{p} \sum_{j=1}^{4} F_{ij}^2 \quad (17)$$

$F_{ij}$ being the generalized force corresponding to the $j$-th actuator when the robot is in the $i$-th configuration. In order to guarantee that the determinant of the Forward Jacobian is different from zero for all configurations considered on the trajectory, the following constraints are imposed,

$$\left|det\left(\Phi_{x_{ref}}\right) - det(\Phi_{x_i})\right| < \left|det\left(\Phi_{x_{ref}}\right)\right|;\ i = 1,2 \dots, p \quad (18)$$

being,

$$det\left(\Phi_{x_{ref}}\right) = max\left(det(\Phi_{x_i})\right);\ i = 1,2 \dots, p \quad (19)$$

The constraints (18) can be written as follows,

$$2 \cdot det\left(\Phi_{x_{ref}}\right) \cdot det(\Phi_{x_i}) - det^2(\Phi_{x_i}) > 0;\ i = 1,2 \dots, p \quad (20)$$

For each of the robot's four actuators, the length between its ends must be such that,

$$\begin{array}{l} l_{min} \le q_{i3} \le l_{max}; i = 1,2,3 \\ l_{min} \le q_{i2} \le l_{max}; i = 4 \end{array} \quad (21)$$

$\alpha_{ji}$ being the angle between by the actuator $j$ in the configuration $i$ with respect to the normal to the mobile platform, the following must be fulfilled,

$$\left|\alpha_{ji}\right| < 45°;\ j = 1,2 \ldots 4;\ i = 1,2 \ldots, p \quad (22)$$

The minimization of the objective function (17) subjected to constraints (20) to (22) constitutes a problem of non-linear optimization with non-linear constraints that could be solved by Quadratic Programming Algorithm with Distributed and Non-Monotone Line Search (NLPQLP) (see [23] and [24]).

### 4.2. Optimization procedure with 4 parameters

To make the reconfiguration of the robot easier and faster, the parameters associated with the mobile platform will be kept constant $(R_m, \beta_{MD}, \beta_{MI})$, leaving the optimization problem reduced to 4 variables. The smaller the number of variable parameters, the more difficult it will be to obtain results, so that in the set of constant parameters a feasible value is sought and away from the extremes through the use of the median. The values assigned to those fixed parameters are taken from the results obtained aby pplying the optimization procedure introduced in Subsection 4.1 to a set of 8 trajectories considered representative of the tasks to be performed by the robot. These trajectories will be introduced in Section 5, and taking the median of these values, the following values have been obtained: $R_m = 0.23\ m$; $\beta_{MD} = 1.59\ rad$; $\beta_{MI} = 1.76\ rad$. By rounded angles in degrees, the values are adjusted to $\beta_{MD} = 90° \approx 1.57\ rad$ and $\beta_{MI} = 100° \approx 1.75\ rad$.

In this case, the variables of the optimization procedure are the parameters associated with the position of the actuators on the fixed platform $(R, ds, \beta_{FD}, \beta_{FI})$.

The following intervals of the variables will be established as follows,

$$\left.\begin{array}{lll} R & \in & [0.20, 0.50]\ m \\ ds & \in & [-0.15, 0.15]\ m \\ \beta_{FD} & \in & [0.1, \pi]\ rad \\ \beta_{FI} & \in & [0.1, \pi]\ rad \end{array}\right\} \quad (23)$$

The objective function is similar to (17) but with four variables:

$$f(R, ds, \beta_{FD}, \beta_{FI}) = \sum_{i=1}^{p} \sum_{j=1}^{4} F_{ij}^2 \quad (24)$$

In this case, the optimization procedure continues to apply the constraints (20) to (22) and the same algorithm is used to solve the problem.

# 5. DESIGN EXAMPLES

Different exercises can be found in the rehabilitation process. In passive exercises, the robot is programmed to follow a specific position reference prescribed by a specialist. Several references have been generated to rehabilitate injured lower limbs. The control unit compares the reference with the robot response, and it establishes the movement control in real-time. The position error obtained is imperceptible, so it indicates that the rehabilitation process has been successful because the exercise indicated by the physiotherapist has been done very accurately.

In Active-resistive exercises, the patient has to overcome a resistance imposed by the physiotherapist. For strengthen plantarflexion, a low-frequency sinusoidal position reference in roll can be used. This exercise aims to keep the platform of the parallel robot in a horizontal position, doing opposed torques (measured by a force sensor) to the motion of this platform.

Finally, Active-assistive exercises are usually done at an early stage in the rehabilitation process when the patient is not able to carry out the required movements by him or herself. In these exercises, the robot assists the patient by calculating the movement references. This reference is calculated in real-time depending on the torque applied by the patient to the force sensor.

A set of 8 test trajectories has been proposed, corresponding to the movements that the robot must execute during the rehabilitation process in passive exercises. All the selected trajectories present some kind of difficulty during their execution. These initial parameters of the robot, listed below, were selected to avoid a trivial singular configuration when the mobile platform was placed horizontally [4].

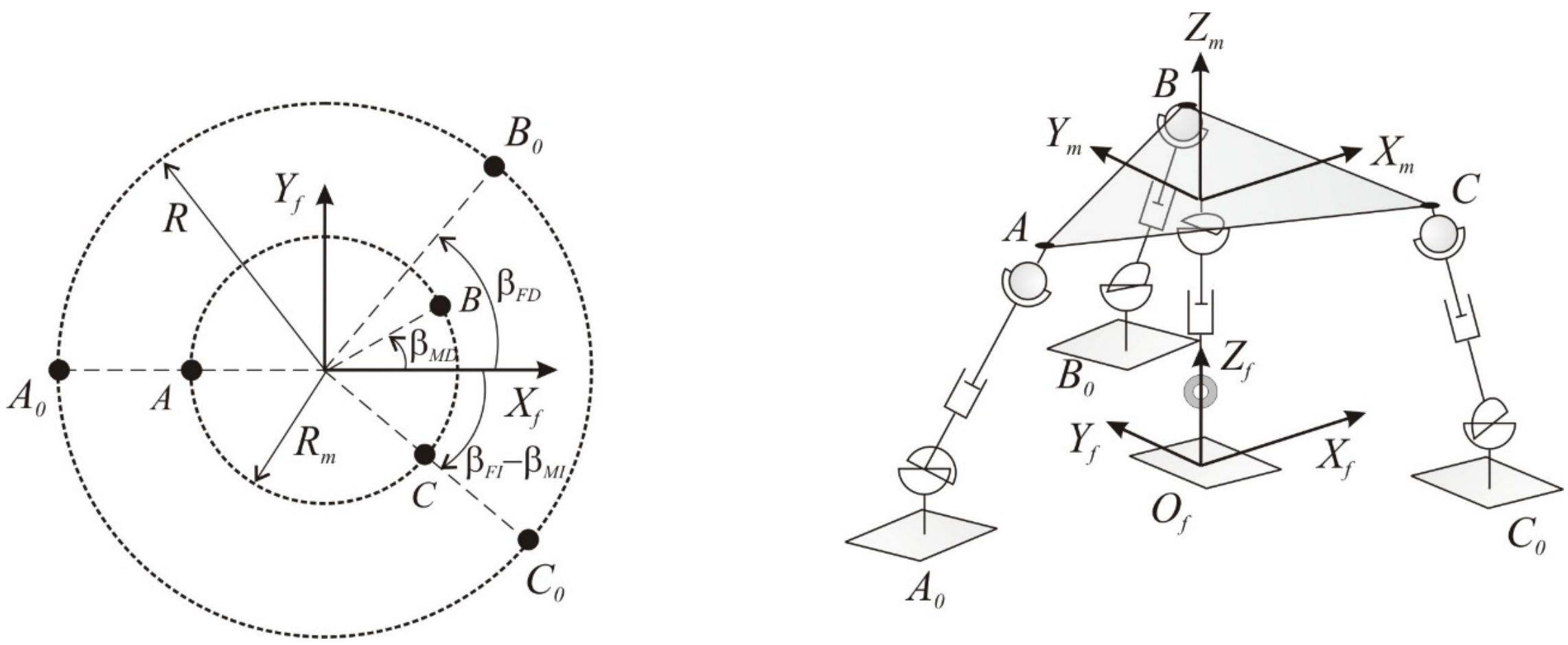

Figure 7. Initial PKM configuration

$$\left.\begin{array}{r} R = 0.4\,m \\ R_m = 0.2\,m \\ ds = 0\,m \\ \beta_{FD} = 50^o \\ \beta_{FI} = 40^o \\ \beta_{MD} = 30^o \\ \beta_{MI} = 40^o \end{array}\right\} \qquad (25)$$

Table 1 shows the main characteristics of the proposed trajectories in terms of the motion of the origin of the mobile reference system and the orientation of the mobile platform: the height of the origin of the mobile reference system could be kept constant, it could experience a vertical displacement of the origin, or it could describe a line or an ellipse in the $X_fZ_f$ plane. The orientation of the mobile platform could be constant or could vary during the movement. In all cases, the modulus of the velocity is constant, as is the order of magnitude that is expected in the actual rehabilitation movements. Likewise, the difficulties found during their execution are reported in the aforementioned table. These difficulties are as follows:

(1) Forward singularities

(2) Actuators out of range

Table 1. Test trajectories

| Trajectory | Horizontal | Vertical | Inclined straight line | Ellipse |
|---|---|---|---|---|
| Constant Orientation | Tr1 (1) | Tr3 (2) | Tr5 (2) | Tr7 (1) and (2) |

| Variable Orientation | Tr2 (1) | Tr4 (1) | Tr6 (1) and (2) | Tr8 (1) and (2) |
|---|---|---|---|---|

To illustrate the aforementioned difficulties, some of the results obtained with trajectory Tr8 will be shown. The prescribed motion for the mobile platform is an ellipse with the following characteristics:

$$x_m = x_0 + v_0 \cdot t$$

$$z_m = z_0 + a \cdot \sqrt{1 - \left(\frac{x}{b}\right)^2}$$

$$\theta = \theta_0 + \omega_\theta \cdot t$$

$$\psi = \psi_0 + \omega_\psi \cdot t$$

$$x_m = x_0 + v_0 \cdot t$$
$$z_m = z_0 + a \cdot \sqrt{1 - \left(\frac{x_m}{b}\right)^2}$$
$$\theta = \theta_0 + \omega_\theta \cdot t$$
$$\psi = \psi_0 + \omega_\psi \cdot t$$

where,

$$x_0 = -0.15\,m; \quad v_0 = 0.015\,m/s; \quad \dot{v}_0 = 0\,m/s^2$$
$$z_0 = 0.25\,m; \quad a = 0.40\,m; \quad b = 0.20\,m$$
$$\theta_0 = 30^\circ; \quad \omega_\theta = -0.05\,rad/s$$
$$\psi_0 = 0^\circ; \quad \omega_\psi = -0.05\,rad/s$$
$$Time = 20.0\,s; \Delta t = 0.30\,s$$

$x_m$, $z_m$ being the coordinates of the origin of the reference system attached to the mobile platform, θ and ψ the Euler angles with regard to the fixed reference system, *t* the time, and $\vec{F}_m = [45.0 \quad 0 \quad -45.0]^T N$ the external forces applied to the mobile platform; no external torque is considered. All the magnitudes are expressed in the local reference system to the mobile platform. First, the inverse kinematic problem in position, velocities, and accelerations is solved for a 20-second motion for 67 robot configurations.

After that, the values obtained for the generalized coordinates, velocities, and accelerations corresponding to the active joints are applied to the forward kinematic problem, obtaining the following results for the 4 coordinates corresponding to the mobile platform (see Figure 8) and for the determinant of the Forward Jacobian (see Figure 9),

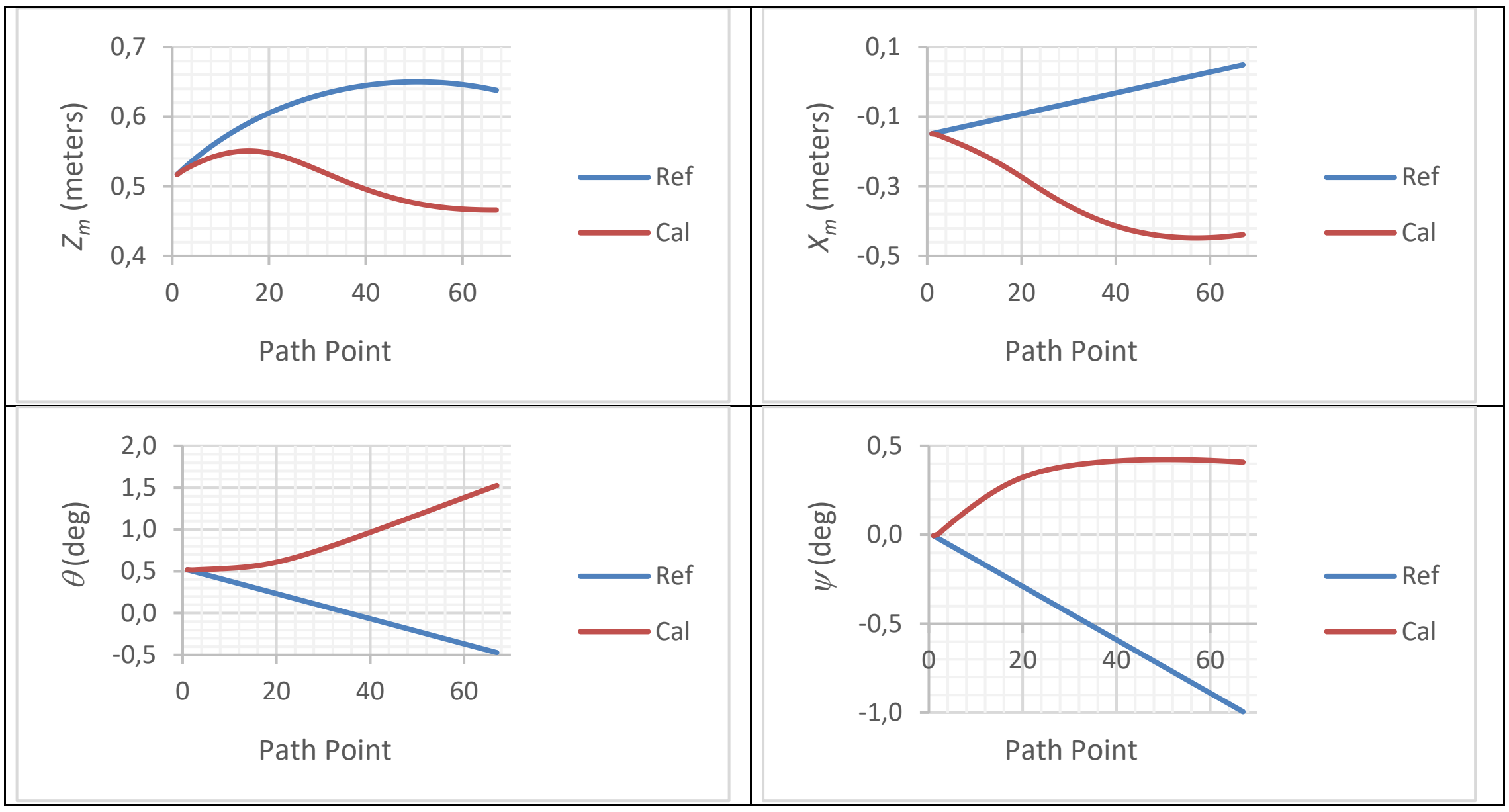

Figure 8. Trajectory Tr8, forward path before reconfiguration.

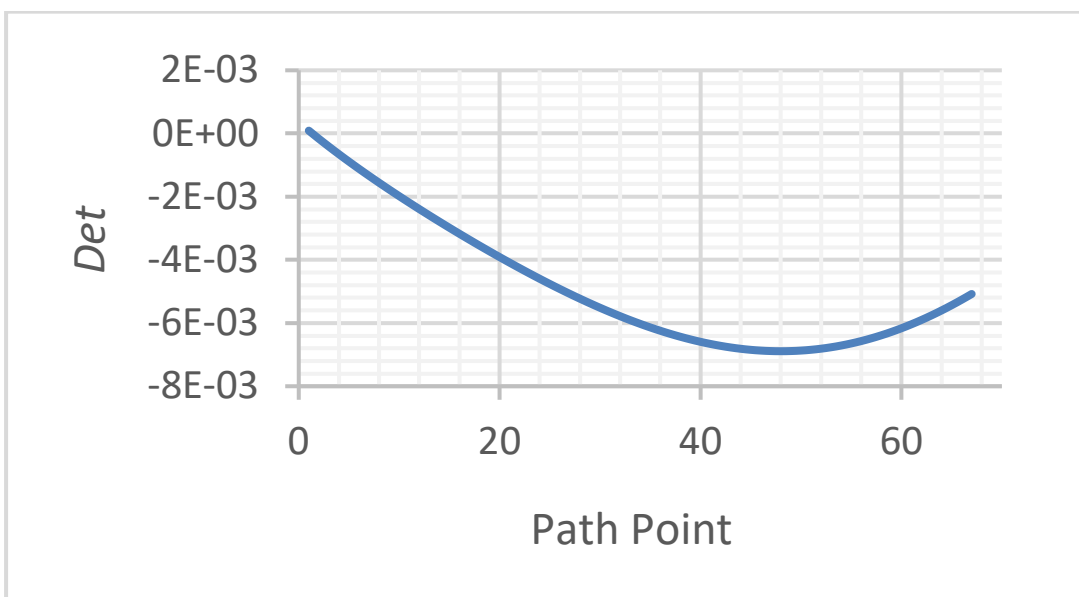

Figure 9. Trajectory Tr8, Determinant of the Forward Jacobian before reconfiguration.

It is remarkable that at the point at which the determinant of the Forward Jacobian becomes null (the manipulator is in a singular configuration), the manipulator adopts a second assembly configuration. In this case, it is solely due to the behavior of the numerical procedure used (Newton-Raphson) to solve the forward kinematic problem of position. However, in the case of the actual robot, the assembly configuration adopted would depend on various considerations (friction, clearances, dynamic behavior) that are very difficult to control and, therefore, unacceptable in a human-mechanical interaction. Finally, the values of the generalized coordinates, velocities, and accelerations are applied to equation (6) to solve the inverse dynamic problem. The active forces needed to perform the prescribed motion and the power consumption are depicted in Figure 10.

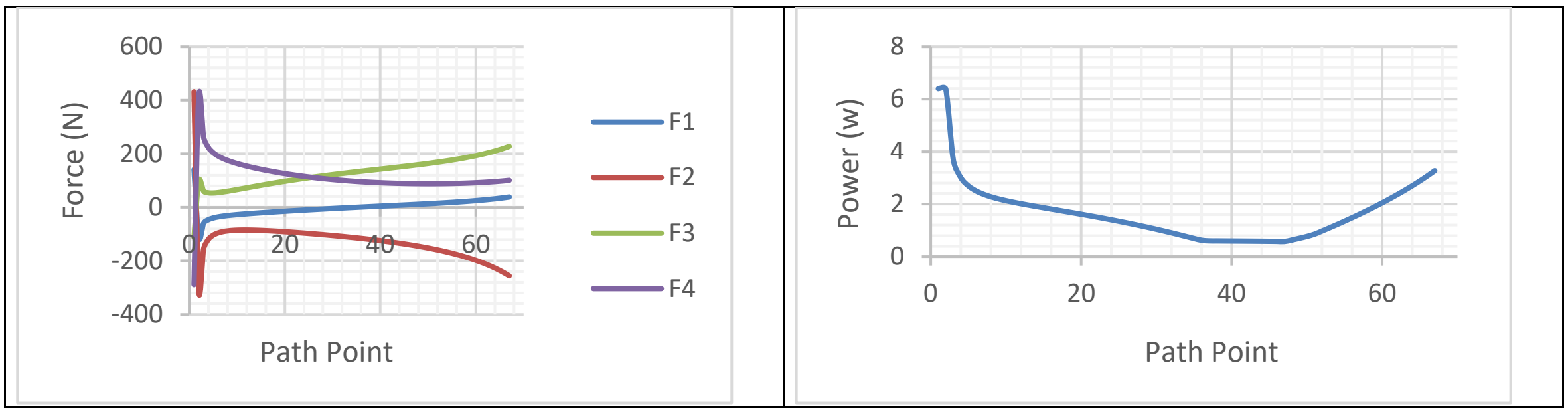

Figure 10. Active forces and power consumption. Before reconfiguration

As can be seen in this figure, the forces (considering the discretization made over the total time of the movement) increase to high values in the vicinity of the forward singularity.

# 6. RESULTS. VIRTUAL ROBOT

The optimization procedure proposed in Section 4 will now be applied to the 8 aforementioned non-feasible trajectories. The initial guess is the initial robot configuration, as described in Section 5.

In Table 2 the values of the objective function (sum of the square values of the 4 active forces on 11 configurations, equations (17) and (24)) are depicted, for the optimization procedure considering 7 and 4 design variables. Obviously, the higher the number of variables, the better the results obtained from the optimization procedure; however, those differences are considered acceptable, bearing in mind that keeping the system as simple as possible is one of the primary goals of the design. Notice that even in the worst-case situation, trajectory Tr7, the active forces needed are feasible.

Table 2. Objective function with 7 and 4 variables.

| | Tr1 | Tr2 | Tr3 | Tr4 | Tr5 | Tr6 | Tr7 | Tr8 |
|---|---|---|---|---|---|---|---|---|
| $F_{7v}$ ($N^2$) | 1.59 $10^5$ | 1.42 $10^5$ | 1.22 $10^5$ | 1.30 $10^5$ | 1.43 $10^5$ | 1.94 $10^5$ | 3.38 $10^5$ | 1.58 $10^5$ |
| $F_{4v}$ ($N^2$) | 1.86 $10^5$ | 1.69 $10^5$ | 1.71 $10^5$ | 1.96 $10^5$ | 1.85 $10^5$ | 2.29 $10^5$ | 19.71 $10^5$ | 7.95 $10^5$ |

Therefore, from this point, only the reconfigurations obtained with the procedure based on the 4 design variables corresponding to the connection of the legs to the fixed platform will be considered. These results are depicted in Table 3.

Table 3. Results of the optimization procedure with four variables

| | Tr1 | Tr2 | Tr3 | Tr4 | Tr5 | Tr6 | Tr7 | Tr8 |
|---|---|---|---|---|---|---|---|---|
| *ds* (*mm*) | 150 | 150 | 102 | 95 | 121 | 150 | -98 | 150 |
| *R* (*mm*) | 242 | 200 | 298 | 381 | 200 | 200 | 200 | 303 |
| $R_m$ (*mm*) | 230 | 230 | 230 | 230 | 230 | 230 | 230 | 230 |
| $\beta_{FD}$ (º) | 170 | 180 | 67 | 90 | 69 | 6 | 72 | 24 |
| $\beta_{FI}$ (º) | 180 | 180 | 161 | 19 | 180 | 180 | 167 | 138 |
| $\beta_{MD}$ (º) | 90 | 90 | 90 | 90 | 90 | 90 | 90 | 90 |
| $\beta_{MI}$ (º) | 100 | 100 | 100 | 100 | 100 | 100 | 100 | 100 |

Next, two of the results obtained will be shown more in detail. The first corresponds to Tr5, which is not feasible with the initial configuration of the robot because one of the actuators fell outside the admissible range, although it did not present problems with forward singularities. The ranges of the actuators will be compared with the original (left) and optimized (right) configuration of the parallel robot (see Figure 11). Likewise, the active generalized forces needed to execute the aforementioned trajectory with the original (left) and optimized (right) configuration will be compared (see Figure 12).

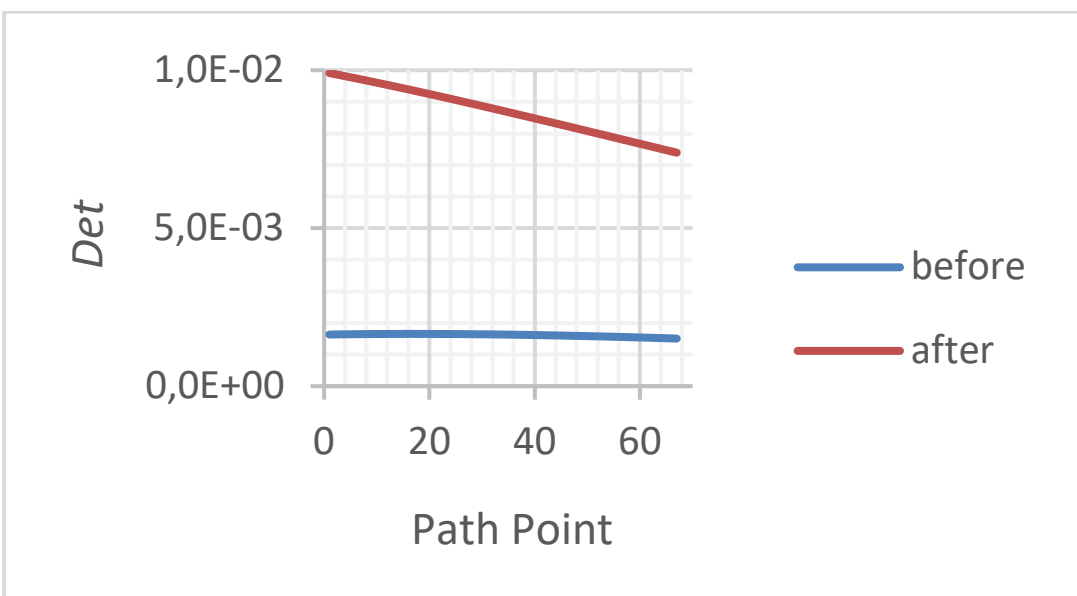

Figure 11. Trajectory Tr5. Determinant of the Forward Jacobian before and after reconfiguration.

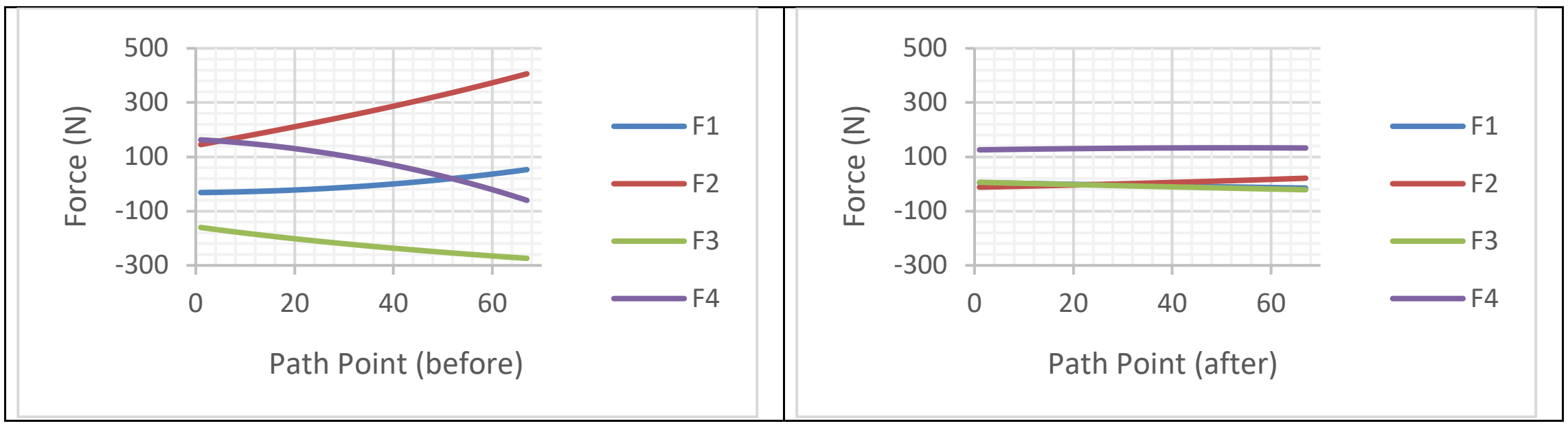

Figure 12 Trajectory Tr5. Active forces before and after reconfiguration.

As shown in Figure 11, not only has the problem with the actuation limits in actuator 1 been corrected but also two of the other actuators (2 and 4) have moved away from their limits. The determinant of the direct Jacobian also moves away from any null or very small values and, in short, the forces with the new configuration of the robot experience an important decrease. The maximum values of forces needed in the actuators are reduced to 29%, 5%, 7%, and 72%, respectively, of their original values (see Figure 12).

As indicated in Section 5 (see Figures 8 and 9), trajectory Tr8 could not be executed properly because it presented configurations where the Forward Jacobian was singular or very close to singularity and because the lengths required by three of its actuators were outside the limits. As shown in Figure 13, these issues have been solved in the new configuration of the RPKM.

Obviously, by eliminating the problem of forward singularity, the high values of generalized forces in their vicinity are reduced, but a decrease in the level of active forces in the rest of the trajectory can also be noted. This is confirmed by comparing the levels of power required before (Figure 10) and after (Figure 14) reconfiguration of the PKM.

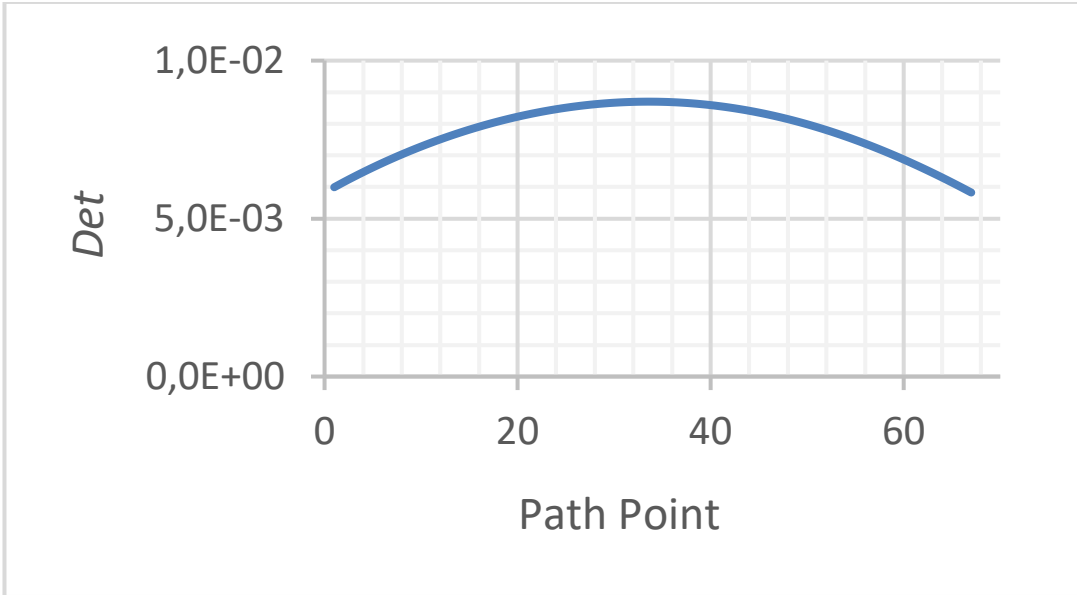

Figure 13. Trajectory Tr8. Determinant of the Forward Jacobian after reconfiguration.

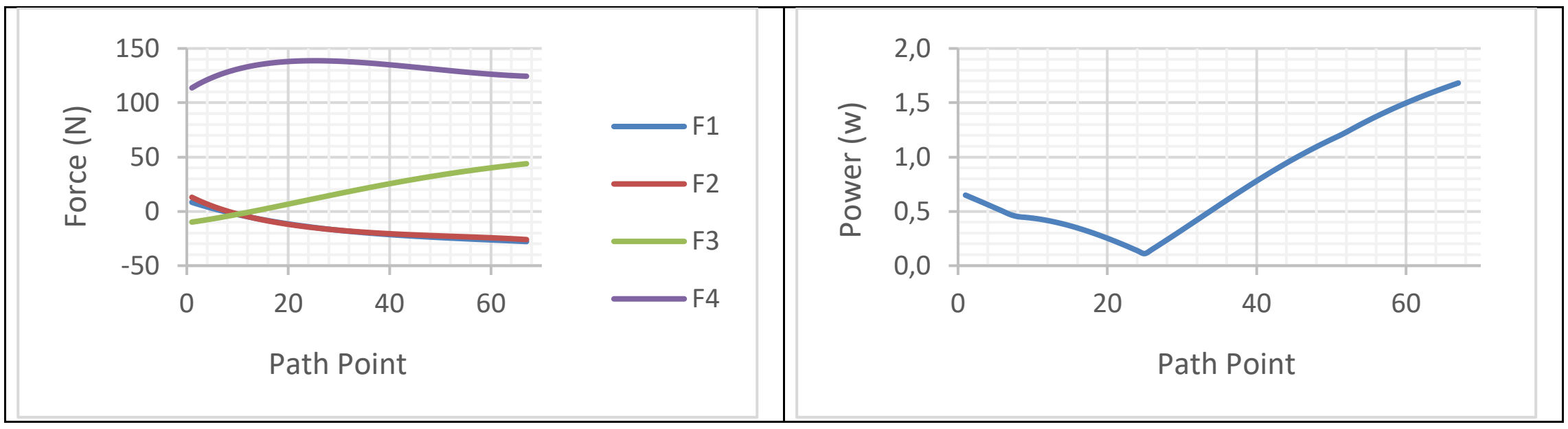

Figure 14. Trajectory Tr8. Active forces and power consumption after reconfiguration.

# 7. RESULTS. ACTUAL ROBOT

## 7.1 Experimental Setup

The problem that arises when a robot operates in the vicinity of a Type II Forward Singularity is that, although the control system can reach the active generalized coordinates that the inverse kinematics establishes for the desired position and orientation of the mobile platform of the PKM in the Cartesian space, it is possible that the robot could adopt another assembly configuration. One way to detect this problem is to measure that position and orientation using an external system. The motion of the PKM mobile platform was therefore captured using stereophotogrammetry (Kinescan, Page, et al. [25]) with a sampling rate of 25 photograms per second. The location of the robot platform was defined using three passive markers. Four markers located on the base of the robot defined the location of the laboratory reference system. The measured standard deviation of the marker coordinate errors was lower than 0.5 mm (see Figure 15).

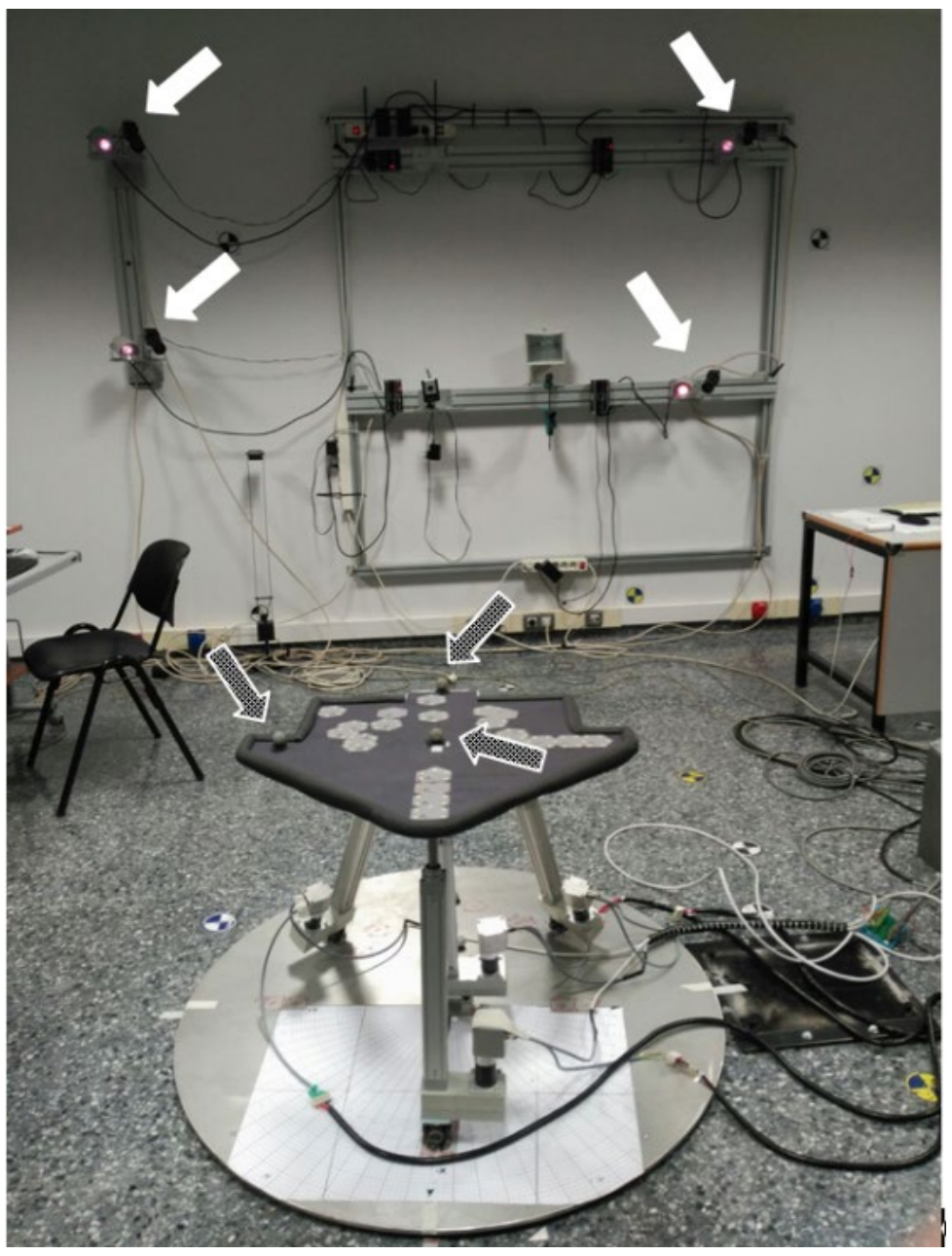

Figure 15. Stereophotogrammetry showing the location of the four cameras and the PKM with three reflective markers attached to the mobile platform

To actuate the actual robot, four Maxon RE40 Graphite Brushes motors have been used. They are compact, powerful, low-inertia 150 Watt motors, and their specifications are 24V nominal voltage, 6940rpm nominal speed, 6A max. continuous current and 2420mNm stall torque.

These actuators are equipped with encoder sensors and brakes. The encoder sensor is the ENC DEDL 9149 system which is a digital incremental encoder with 500 pulses per revolution, 3 channels, and 100 kHz max. operating frequency. The brake system is the Brake AB 28 system, which is a 24 V, 0.4 Nm permanent-magnet, single-face brake for DC motors that prevents rotation of the shaft at a standstill or when the motor power is turned off.

To develop the control architecture for the parallel robot, an industrial PC has been used. It is based on a high-performance 4U Rackmount industrial system with 7 PCI slots and 7 ISA slots. The industrial PC is equipped with a 2.5 GHz Intel® Pentium® Core 2 Quad/Duo processor, 4 GB SDRAM, and two Advantech™ data acquisition cards: PCI-1720 and PCL-833. The first card provides the Digital/Analog conversions and is used to supply the control actions for each parallel robot actuator. The second card is used to read the encoder signal to measure the positions of the four prismatic joints of the robot.

The industrial PC is also equipped with the Linux Ubuntu system, the real-time, component-based middleware Open Robot Control Software (OROCOS), and the high-level middleware Robot Operating System (ROS). These are all free, open software, so this control architecture has several advantages: it is a very economical, totally open system and it enables different control strategies to be implemented using different sensors, such as potentiometers, force sensors, and machine vision cameras, to name but a few.

### 7.2 Results

To validate the proposed procedure, two robot configurations will be considered, the original one and the one optimized for the specific trajectory Tr1 (see Figure 16). The prescribed trajectory will be discretized in the Cartesian space at 150 intermediate points, and the inverse problem of position will be solved for both configurations.

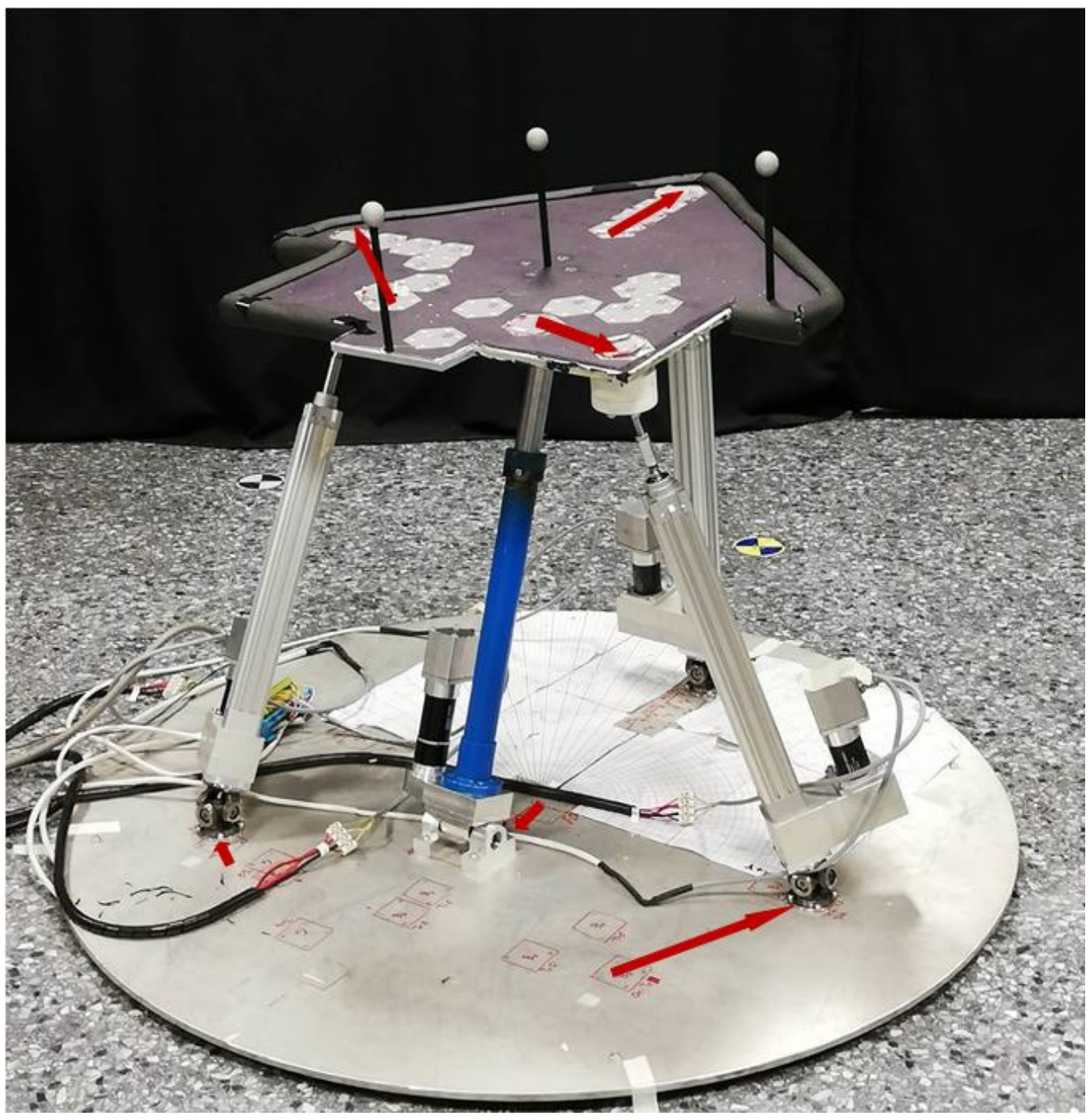

Figure 16. Reconfiguration from the original one to Tr1

The trajectory Tr1, considered to perform a test on the actual robot, is a horizontal displacement, at a constant velocity, of the origin of the mobile reference system, keeping the orientation of said platform constant. The coordinates ($x_m$, $z_m$) of the origin of the mobile reference system in its initial position are (-0.048, 0.631) $m$.

Figure 17.1 presents the Cartesian reference and the robot response. It shows the motion reference (in blue), the robot response for the original configuration (in black), and the executed trajectory of the actual robot for the Tr1 configuration (in red). This response was obtained using the cameras of the stereophotogrammetry system by recording the actual robot's movements. The reference consists of a first movement in the Z-axis (from 0.631m to 0.72m). In the next 5 seconds, the robot changes the rotation angles $\theta$ and $\psi$ of the mobile platform from the original orientation to 0 radians. After that, a linear movement is executed in the X-axis from -0.048m to 0.152m at a constant velocity of 0.02m/s for 10 seconds. The robot remains in this position for 5 seconds, before performing the inverse movements to return to the origin. A video with the actual robot performing this trajectory can be found at: (https://imbio3r.ai2.upv.es/nuevo_video/actual-robot-stereophotogrammetry-validation).

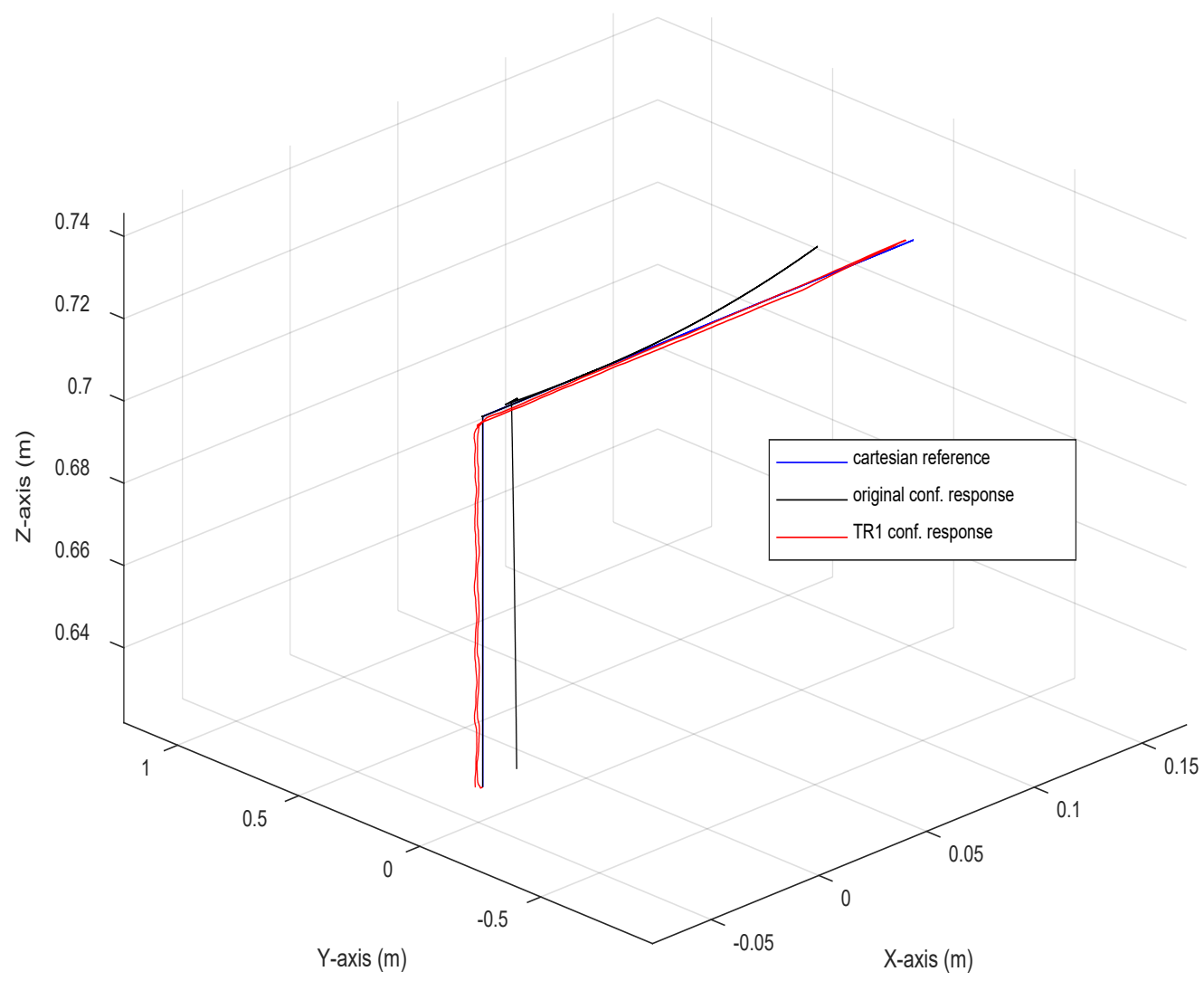

17.1

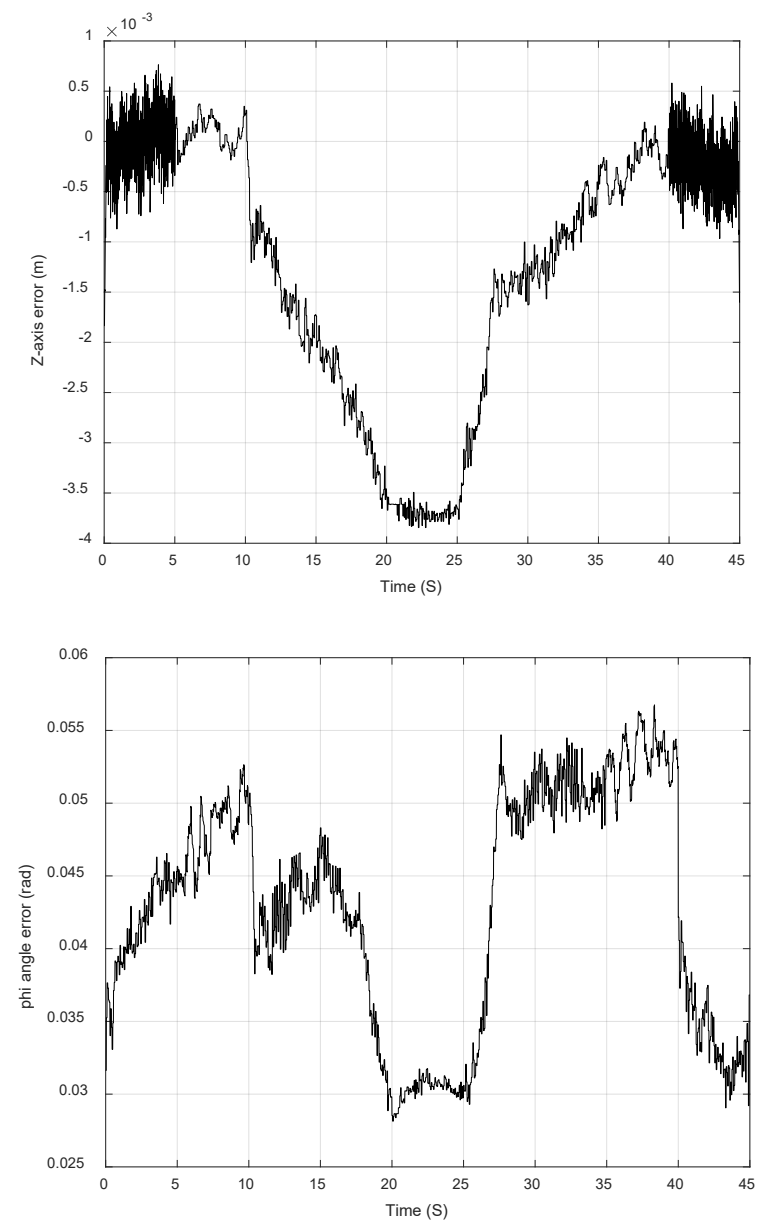

17.2

Figure 17. Executed 3D trajectory and errors of the actual PKM

As can be seen in Figure 17, the robot with the Tr1 configuration can follow the reference without problems, providing a small error in the translation ($X$ and $Z$) and rotation (theta, phi) axes. However, the robot with the original configuration passes through a forward singularity at

t=12.5 seconds (approx.), so it is impossible to keep following the movement reference and its response is dramatically worse.

Figure 18 presents the response for the actual parallel robot. Figure 18.1 shows the position reference and the active coordinate of limb1 to execute the prescribed trajectory. Figure 18.2 presents the control actions applied to the actuator. Notice that the robot's coordinate follows the reference very accurately, and the control actions are maintained at a low level throughout the trajectory.

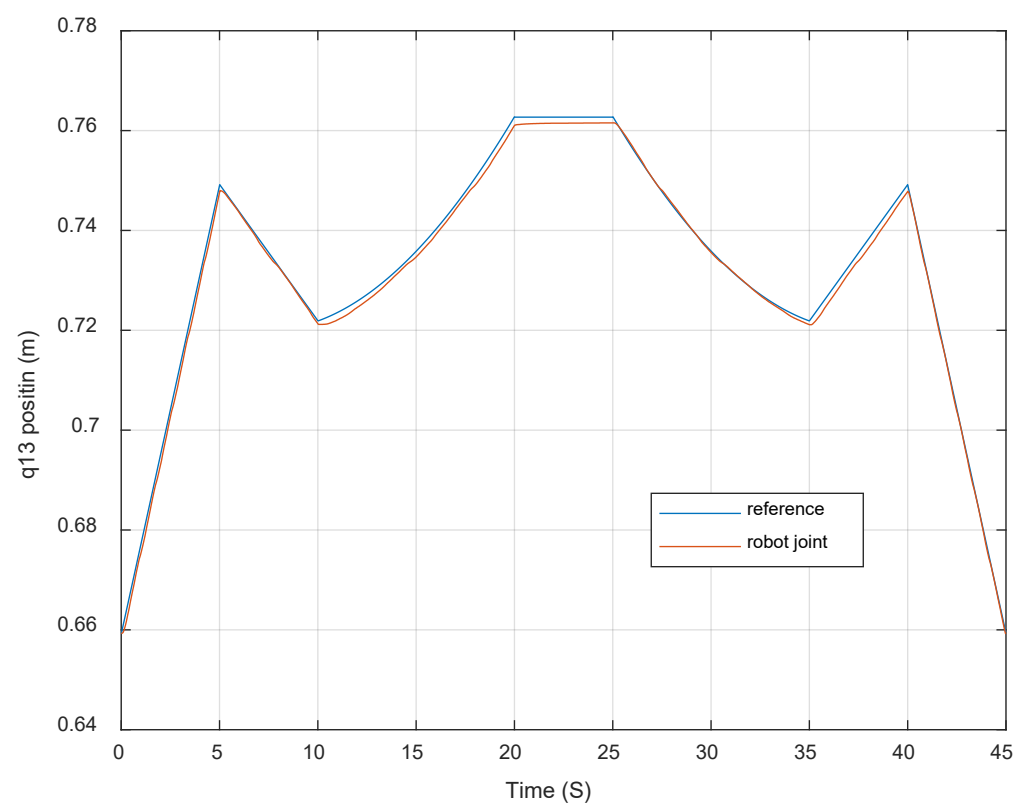

18.1

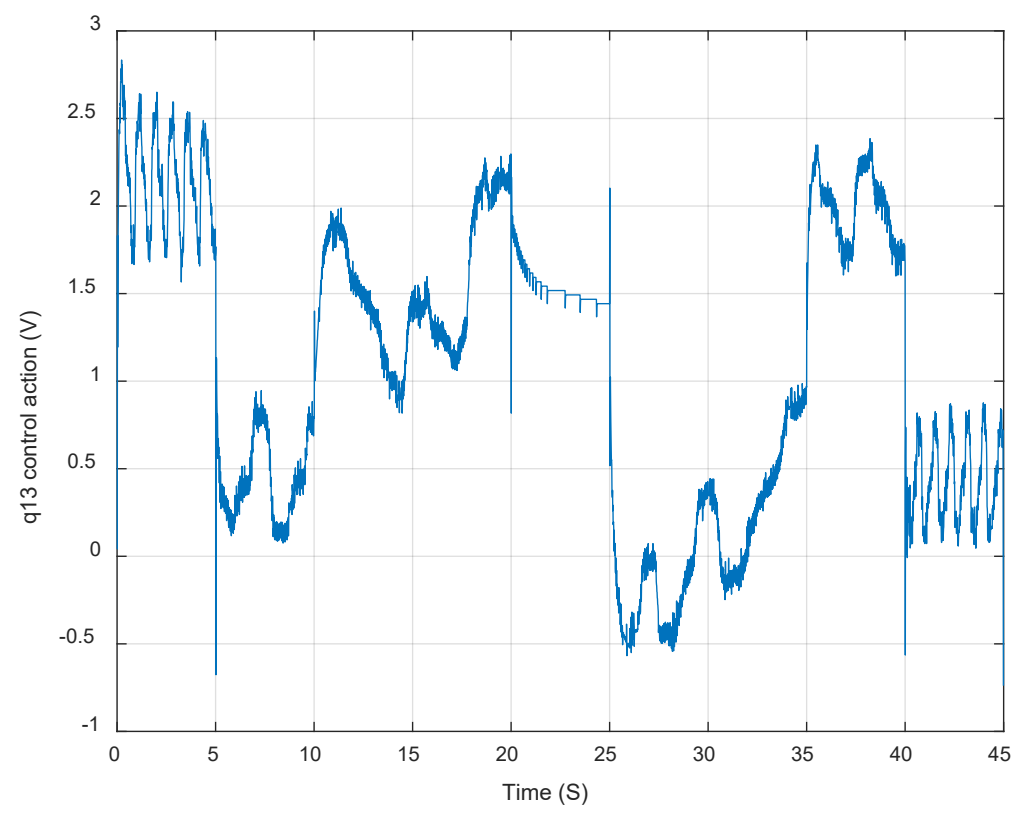

18.2

Figure 18. First limb response for the actual PKM

# 8. CONCLUSIONS

The four degrees of freedom 3UPS-RPU PKM is able to perform the necessary movements for diagnosis and rehabilitation tasks on human knees, particularly in conditions affecting the anterior cruciate ligaments. However, it has been proven that for the original design of the robot, the execution of certain rehabilitation trajectories is not possible because: a) the Forward Jacobian becomes singular, which gives rise to control problems and an increase in the forces needed to execute said trajectories, and b) the values required by the active generalized coordinates fall outside the operating range of the prismatic actuators. Given the application that is sought, it is not possible to modify the trajectories to avoid singularities, which are identified as the main problem during the operation of the PKM; thus, a reconfiguration strategy of the PKM has been considered by modifying the attachment points of the legs on both the fixed and mobile platforms of the robot.

The reconfiguration has been proposed as a non-linear optimization problem subject to non-linear constraints. The objective function to be minimized is the sum of the square of the active generalized forces on a set of selected via points on the trajectory to be performed. As the constraint to verify, we have a non-singular Forward Jacobian in the aforementioned points and the active generalized coordinates must be within the admissible values for the PKM actuators. For this, a dynamic inverse model of the robot has been obtained using D'Alembert's Principle and the Principle of the Virtual Powers. The optimization problem has been solved using a Sequential Quadratic Programming algorithm, using as an initial guess the original configuration of the robot. A set of eight non-feasible trajectories has been considered and a feasible solution to the optimization problem has been found for all of them.

For the eight trajectories, a comparison was made of the results obtained with the reconfigurations of 7 variables (modifying parameters in both platforms) and 4 variables (only in the fixed platform). From the results obtained it was concluded that, to keep the mechanical complexity and the cost of the robot as low as possible, it was enough to base the reconfiguration only on the modification of the parameters corresponding to the connection of the legs to the fixed platform.

The new configurations obtained from the optimization process have been tested on a virtual model and an actual one. In all cases, it has been proven that only a new sequence of configurations does not pass through the forward singularity and that the active generalized coordinates are within the physical ranges admissible by the actuators. In addition, the forces needed to perform the trajectories are much lower than those required under the initial configuration of the robot.

ACKNOWLEDGEMENTS

This work was supported by the Spanish Ministry of Education, Culture, and Sports through the Project for Research and Technological Development with ref. DPI2017-84201-R

The authors would like to thank Prof. K. Schittkowski for providing the source code for solving nonlinearly constrained optimization problems.